\documentclass[10pt,twocolumn,letterpaper]{article}

\usepackage{cvpr}              
\definecolor{cvprblue}{rgb}{0.21,0.49,0.74}
\usepackage[pagebackref,breaklinks,colorlinks,allcolors=cvprblue]{hyperref}
\usepackage{multirow}

\def\paperID{} 
\def\confName{CVPR}
\def\confYear{2026}

\title{Evo-1: Lightweight Vision-Language-Action Model \\
with Preserved Semantic Alignment}

\author{
Tao Lin$^{1,2,3}$ \quad Yilei Zhong$^{1}$ \quad Yuxin Du$^{1,2,3}$ \quad Jingjing Zhang$^{1}$ \quad Jiting Liu$^{1}$  \quad Yinxinyu Chen$^{5}$ \\
\quad Encheng Gu$^{6}$ \quad Ziyan Liu$^{1}$ \quad Hongyi Cai$^{1}$ \quad Yanwen Zou$^{1,3,4}$ \quad Lixing Zou$^{1}$ \quad Zhaoye Zhou$^{1}$ \\  \quad Gen Li$^{1,3,7\dagger}$  \quad Bo Zhao$^{1,2,3\dagger}$\\
$^{1}$School of AI, Shanghai Jiao Tong University \quad
$^{2}$EvoMind Tech \quad
$^{3}$IAAR-Shanghai  \quad
$^{4}$SII  \quad
\\
$^{5}$Carnegie Mellon University \quad
$^{6}$University of Cambridge\quad
$^{7}$Nanyang Technological University 
\\
\texttt{taolin200108@gmail.com, bo.zhao@sjtu.edu.cn} \\
\url{https://github.com/MINT-SJTU/Evo-1}
}

\makeatletter
\newcommand\blfootnote[1]{%
  \begingroup
  \renewcommand\thefootnote{}\footnote{#1}%
  \addtocounter{footnote}{-1}%
  \endgroup
}
\makeatother

\usepackage{booktabs,tabularx,array,multirow}
\renewcommand{\arraystretch}{0.9}
\newcolumntype{L}[1]{>{\raggedright\arraybackslash}m{#1}}

\begin{document}
\maketitle
\blfootnote{$\dagger$~Corresponding authors.}

\begin{abstract}
Vision-Language-Action (VLA) models have emerged as a powerful framework that unifies perception, language, and control, enabling robots to perform diverse tasks through multimodal understanding. 
However, current VLA models typically contain massive parameters and rely heavily on large-scale robot data pretraining, leading to high computational costs during training, as well as limited deployability for real-time inference.
Moreover, most training paradigms often degrade the perceptual representations of the vision-language backbone, resulting in overfitting and poor generalization to downstream tasks.
In this work, we present \textbf{Evo-1}, a lightweight VLA model that reduces computation and improves deployment efficiency, while maintaining strong performance without pretraining on robot data. 
Evo-1 builds on a native multimodal Vision-Language model (VLM), incorporating a novel cross-modulated diffusion transformer along with an optimized integration module, together forming an effective architecture.
We further introduce a two-stage training paradigm that progressively aligns action with perception, preserving the representations of the VLM.
Notably, with only \textbf{0.77 billion} parameters, Evo-1 achieves \textbf{state-of-the-art} results on the Meta-World and RoboTwin suite, surpassing the previous best models by 12.4\% and 6.9\%, respectively, and also attains a competitive result of 94.8\% on LIBERO.
In real-world evaluations, Evo-1 attains a 78\% success rate with high inference frequency and low memory overhead, outperforming all baseline methods.
We release code, data, and model weights to facilitate future research on lightweight and efficient VLA models.
\end{abstract}    
\section{Introduction}
\label{sec:intro}

In recent years, Vision-Language models (VLMs) \cite{alayrac2022flamingo,wang2024qwen2,zhu2025internvl3,achiam2023gpt} have achieved remarkable progress in multimodal understanding and reasoning.
Inspired by these advances, researchers have extended multimodal learning to robotic control, leading to the development of Vision-Language-Action (VLA) models \cite{kim2024openvla,black2410pi0,zitkovich2023rt,lin2025evo,bjorck2025gr00t}.
VLA models integrate perception, language, and control, enabling robots to follow natural language instructions grounded in visual observations and perform diverse manipulation tasks with strong generalization across environments and embodiments.

Despite their promising capabilities, existing VLA models face several critical limitations.
First, their massive number of parameters, often reaching several billions, leads to substantial GPU memory usage and high computational costs during both training and inference.
Second, their large computational overhead leads to a low control frequency, limiting the model’s real-time responsiveness in interactive robotic tasks.
Third, the widely adopted end-to-end training paradigm often degrades the representation space of the vision-language backbone, leading to poor generalization and overfitting in downstream tasks.
Fourth, the majority of these models strongly rely on long-duration training over large-scale robot datasets (e.g., OXE~\cite{o2024open}, DROID~\cite{khazatsky2024droid}), whose collection is labor-intensive and costly. 

In this work, we introduce Evo-1, a lightweight VLA model designed for low-cost training and real-time deployment.
Evo-1 adopts a unified vision-language backbone~\cite{zhu2025internvl3}
pretrained under a single-stage multimodal paradigm, where perceptual and linguistic representations are learned jointly without post-hoc alignment, enabling strong multimodal perception and understanding. This compact VLM design substantially reduces overall model scale, reducing GPU memory requirements and computational demands in both training and inference.
On top of this backbone, we design a cross-modulated diffusion transformer that models continuous action trajectories, allowing efficient temporal reasoning for consistent motion generation. This design also contributes to the model's compactness and greatly increases inference frequency, supporting responsive behavior in real-time interactive robotic scenarios. We further introduce an optimized integration module that aligns the fused vision-language representations with the proprioceptive information of robot, thereby enabling seamless incorporation of multimodal features into the subsequent control.
To strike a balance between preserving the inherent multimodal representational capacity and enabling effective adaptation to downstream action generation, we propose a two-stage training paradigm that gradually aligns the perception and control modules while substantially mitigating distortion of the VLM's semantic space. By preserving the inherited semantic space, the model demonstrates strong generalization and competitive results without robot data pretraining.

Evo‑1 achieves strong results across three challenging simulation benchmarks: it sets a new state-of-the-art on Meta‑World (80.6\%) and RoboTwin suite (37.8\%), surpassing previous bests of 68.2\% and 30.9\%, respectively, and reaches 94.8\% on LIBERO, demonstrating its adaptability in both single-arm and dual-arm manipulation tasks.
In real-world evaluations on four representative robotic tasks, Evo-1 achieves an overall success rate of 78\%, consistently outperforming other baselines. It also delivers high inference frequency with a compact GPU memory utilization, demonstrating both computational efficiency and stable control in physical deployments.
Our contributions are summarized as follows:

\begin{enumerate}
    \item \textbf{Lightweight and efficient architecture.} We propose Evo-1, a lightweight VLA architecture with only 0.77B parameters that reduces training cost and improves inference speed for real-time deployment on consumer-grade GPUs.
        
    \item \textbf{Semantic preservation for improved generalization.} We introduce a two-stage training paradigm that strikes a balance between preserving inherent multimodal understanding of the VLM and adapting it to downstream action generation, effectively enhancing generalization across diverse manipulation tasks.
    
    \item \textbf{Strong performance without pretraining.} Extensive experiments in both simulation and real-world tasks demonstrate that Evo-1 achieves state-of-the-art performance without relying on large-scale robot data pretraining, substantially reducing the need for costly and labor-intensive data collection.
\end{enumerate}
\section{Related Work}
\label{sec:related_work}

\noindent \textbf{Large-Scale Vision-Language-Action Models.}
Recent research has advanced Vision-Language-Action (VLA) models \cite{black2410pi0,bjorck2025gr00t,kim2024openvla,lin2025evo,shukor2025smolvla,wen2025tinyvla,liu2025hybridvla} that integrate perception, language, and control within a unified multimodal framework. These models extend pre-trained vision-language backbones \cite{alayrac2022flamingo,beyer2024paligemma,zhu2025internvl3,biderman2023pythia,marafioti2025smolvlm} to predict robot actions, enabling impressive few-shot generalization across diverse manipulation tasks \cite{li2025learning,mon2025embodied}.
Representative works such as OpenVLA \cite{kim2024openvla} utilize large-scale demonstration data from the Open-X Embodiment dataset \cite{o2024open}, achieving cross-embodiment transfer through discrete action modeling. $\pi_0$ \cite{black2410pi0} adapts the PaliGemma \cite{beyer2024paligemma} architecture with a flow-matching-based action expert for continuous control, while Hi-Robot \cite{shi2025hi} introduces hierarchical reasoning and dual-expert architectures for long-horizon planning.

Although these models demonstrate remarkable performance and generalization, they commonly rely on large pretrained backbones with billions of parameters, leading to significant computational demands and limited feasibility for real-time robotic deployment.

\noindent \textbf{Lightweight and Efficient Vision-Language-Action Models.} While large-scale VLA models achieve strong generalization, their substantial computational costs hinder practical deployment.
To improve efficiency, recent studies \cite{wen2025tinyvla,shukor2025smolvla,wang2025vla} have explored compact architectures that retain multimodal reasoning with significantly fewer parameters.
TinyVLA \cite{wen2025tinyvla} proposes a sub-billion-parameter VLA framework that combines lightweight vision-language backbone with a diffusion-based policy decoder. SmolVLA \cite{shukor2025smolvla} further emphasizes accessibility by employing a SmolVLM-2 \cite{marafioti2025smolvlm}  backbone and a compact flow-matching action expert, together with layer skipping, token reduction, and asynchronous inference.
Although both models significantly improve efficiency and accessibility, their overall task performance and robustness remain less satisfactory in complex manipulation settings.

Sharing the same goal of advancing efficient VLA modeling, our proposed Evo-1 further contributes to the development of lightweight yet effective architectures that eliminate large-scale pretraining while substantially reducing training cost, inference resource consumption, and deployment complexity, achieving strong and reliable performance across diverse robotic tasks.

\section{Method}
\label{sec:method}

\begin{figure*}[!t]
  \centering
  \includegraphics[width=\textwidth]{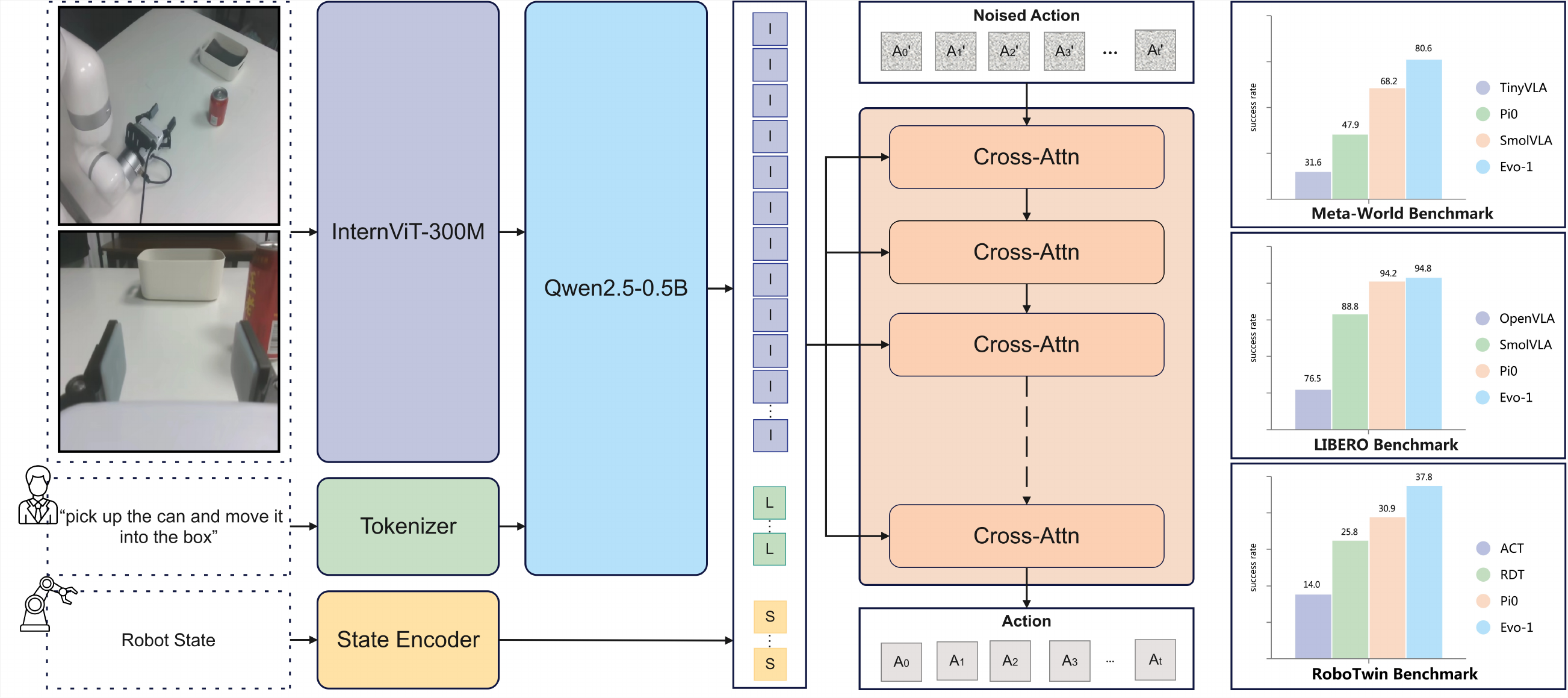} 
\caption{\textbf{Architecture of Evo-1.} The input RGB observations and language instructions are first encoded by a compact vision-language backbone. 
Their fused representations are aligned with the robot state through an optimized integration module and then processed by a cross-modulated diffusion transformer to generate actions. 
The right side shows results across three simulation benchmarks.}
  \label{fig:archi}
\end{figure*}

\subsection{Overview of Evo-1 Architecture}

Evo-1 adopts a modular Vision-Language-Action (VLA) architecture that integrates perception, reasoning, and control within a unified yet computationally efficient framework.
As illustrated in Figure~\ref{fig:archi}, the architecture comprises three core components:
(1) a \textbf{vision-language backbone} that encodes multimodal representations from visual observations and textual instructions;
(2) a \textbf{cross-modulated diffusion transformer} that generates continuous control actions; and
(3) an \textbf{integration module} that bridges perception and control through efficient alignment of multimodal and proprioceptive representations.

Together, these components form a unified perception-language-action pipeline.
Given a set of multi-view visual inputs $\{ I_t^i \}_{i=1}^{N}$, a language instruction $L_t$, and the robot state $s_t$,
the vision-language backbone produces multimodal representations that are propagated through the integration module and interact with the cross-modulated diffusion transformer to produce the final control output.
The overall mapping can be expressed as:
\begin{equation}
a_t = f_{\text{Evo-1}}\!\left(\{ I_t^i \}_{i=1}^{N}, L_t, s_t; \theta \right),
\end{equation}
where $a_t \in \mathbb{R}^{d_a}$ denotes the continuous action vector executed at time $t$, and $\theta$ represents the learnable parameters of the entire model.
This formulation summarizes the end-to-end process of Evo-1, effectively bridging high-level semantic understanding and low-level motor control within a lightweight and computationally efficient framework.

\subsection{Model Design}
\subsubsection{Vision-Language Backbone}
Evo-1 employs the InternVL3-1B model~\cite{zhu2025internvl3} as its vision-language backbone, which was pretrained under a native multimodal paradigm. 
Unlike post-hoc alignment pipelines that retrofit text-only LLMs to handle images, InternVL3 jointly learns linguistic and visual understanding from large-scale multimodal and textual corpora, enabling tight cross-modal alignment and efficient feature fusion.

The visual encoder adopts InternViT-300M~\cite{gao2024mini}, a lightweight transformer distilled from InternViT-6B through layer-wise negative cosine similarity loss. 
Each RGB observation $\{ I_t^i \}_{i=1}^{N}$ is resized to $448\times448$ and passed through a pixel-unshuffle downsampling operation, reducing the number of visual tokens by $4{\times}$. 
This yields compact yet expressive patch embeddings that preserve spatial granularity and maintain generalization across diverse visual domains. 

The language branch leverages Qwen2.5-0.5B~\cite{bai2025qwen2}, a transformer-based decoder with 0.5B parameters. 
Despite its small size, it demonstrates strong capability in capturing diverse task semantics, including spatial, logical, and temporal relations from the instruction $L_t$.

For vision-language fusion, InternVL3-1B inserts patch-level image embeddings into the token sequence by replacing a designated \texttt{<img>} placeholder token.
The resulting fused sequence is processed by the shared transformer decoder, enabling joint reasoning over visual and linguistic context in a unified embedding space.

The fused representation produced by the backbone is denoted as
\begin{equation}
z_t = f_{\text{VLM}}\!\left(\{ I_t^i \}_{i=1}^{N}, L_t\right),
\end{equation}
where $z_t \in \mathbb{R}^{d_z}$ denotes the fused multimodal representation that jointly encodes visual and linguistic information, serving as the input to the integration  module.
To better adapt the pretrained VLM to embodied visuomotor tasks, we retain only the first 14 layers of the language branch, as intermediate layers have been empirically found to exhibit stronger cross-modal alignment between visual and linguistic features~\cite{shukor2025smolvla}, making them more effective for visuomotor control.

\subsubsection{Cross-modulated Diffusion Transformer}

Evo-1 adopts a conditional denoising module as action expert to predict continuous control actions from the fused multimodal embedding produced by the vision-language backbone. Following the flow-matching paradigm~\cite{lipman2022flow, liu2022rectified}, it learns a time-dependent vector field that progressively transforms an initial noisy action into the ground-truth target.

Specifically, the action expert is implemented as a Diffusion Transformer (DiT)~\cite{peebles2023scalable} that solely relies on stacked cross-attention layers, in contrast to the alternating self-attention and cross-attention structure adopted by prior VLA models~\cite{black2410pi0,shukor2025smolvla}.  
Each noisy action sequence $A_t^{\tau}$ is generated by linearly interpolating between the ground-truth action $A_t$ and a randomly sampled noise vector $\epsilon$:
\begin{equation}
A_t^{\tau} = \tau A_t + (1{-}\tau)\epsilon.
\end{equation}
The interpolation weight $\tau$ is sampled from a Beta distribution and clamped to the range $[0.02, 0.98]$ to ensure numerical stability during training.  

During training, the action expert is optimized to learn a time-conditioned velocity field $\mathbf{v}_{\theta}$ that drives the interpolated action $A_t^{\tau}$ toward the ground-truth action $A_t$ under the multimodal context $z_t$ and robot state $s_t$.  
The objective follows the flow-matching formulation~\cite{lipman2022flow,liu2022rectified}, defined as:
\begin{equation}
\resizebox{0.90\linewidth}{!}{$
\mathcal{L}^{\tau}(\theta) =
\mathbb{E}_{p(A_t|z_t,s_t),\,q(A_t^{\tau}|A_t)}\!\left[
\left\|
\mathbf{v}_{\theta}(A_t^{\tau}, z_t, s_t)
- \mathbf{u}(A_t^{\tau} \mid A_t)
\right\|^2
\right],
$}
\label{eq:flowmatching}
\end{equation}
where $\mathbf{u}(A_t^{\tau} \mid A_t)$ denotes the target flow direction that guides $A_t^{\tau}$ toward $A_t$.

At inference time, the final action trunk $\hat{A}_t = [\hat{a}_t, \hat{a}_{t+1}, \dots, \hat{a}_{t+H-1}]$ is predicted by the action expert, conditioned on the fused representation $z_t$, the current robot state $s_t$, and the interpolated action $A_t^\tau$.
\begin{equation}
\hat{A}_t = f_{\text{AE}}(z_t, s_t, A_t^\tau),
\label{eq:action_expert}
\end{equation}
where $f_{\text{AE}}$ denotes the conditioned action expert network that generates a sequence of $H$ future actions aiming to approximate the ground-truth action sequence $A_t$.

\subsubsection{Integration Module}

Evo-1 adopts a cross-attention-based integration module to effectively fuse multimodal and proprioceptive information before conditioning the Cross-modulated Diffusion Transformer.  
The fused multimodal representation $z_t$ is extracted from the 14\textsuperscript{th} layer of the vision-language backbone, capturing intermediate-level semantics that balance visual and linguistic features.  
To preserve the complete information from both the perceptual embedding and the robot’s proprioceptive state, we concatenate $z_t$ with the robot state $s_t$ instead of projecting them into a shared embedding space.  
This concatenated feature serves as the key-value input for the transformer blocks of the action expert, providing a global and information-preserving context for action generation.  
Additional integration variants and their comparative results are detailed in the ablation studies (Sec.~\ref{sec:ablation}).

\begin{table*}[t]
\centering
\setlength{\tabcolsep}{5pt}
\renewcommand{\arraystretch}{0.85}
\resizebox{\linewidth}{!}{
\begin{tabular}{llcccccccc}
\toprule
\textbf{Benchmark} & \textbf{Models} & \textbf{Params}  & \textbf{Robo-Pretrain} 
& \multicolumn{6}{c}{\textbf{Success Rate (\%)}} \\
\midrule
\textbf{Meta-World} & & & & \textbf{Easy} & \textbf{Medium} & \textbf{Hard} & \textbf{Very Hard} & \textbf{Avg.}  \\
\midrule
& Diffusion Policy~\cite{chi2023diffusion} & - & No & 23.1 & 10.7 & 1.9 & 6.1 & 10.5  \\
& TinyVLA-H~\cite{wen2025tinyvla} & 1.3B  & No & 77.6 & 21.5 & 11.4 & 15.8 & 31.6  \\
& $\pi_0$~\cite{black2410pi0} & 3.5B  & Yes & 71.8 & 48.2 & 41.7 & 30.0 & 47.9  \\
& SmolVLA~\cite{shukor2025smolvla} & 2.25B  & No & \underline{87.1} & \underline{51.8} & \underline{70.0} & \underline{64.0} & \underline{68.2}  \\
\midrule
& \textbf{Evo-1 (Ours)} & 0.77B  & No & \textbf{89.2} & \textbf{76.8} & \textbf{77.2} & \textbf{79.2} & \textbf{80.6}  \\
\midrule
\textbf{LIBERO} & & & & \textbf{Spatial} & \textbf{Object} & \textbf{Goal} & \textbf{Long} & \textbf{Avg.}  \\
\midrule
& OpenVLA~\cite{kim2024openvla} & 7B  & Yes & 84.7 & 88.4 & 79.2 & 53.7 & 76.5 \\
& CoT-VLA~\cite{zhao2025cot} & 7B  & Yes & 87.5 & 91.6 & 87.6 & 69.0 & 81.1 \\
& $\pi_0$-FAST~\cite{pertsch2025fast} & 3.5B  & Yes & \underline{96.4} & 96.8 & 88.6 & 60.2 & 85.5  \\
& SmolVLA~\cite{shukor2025smolvla} & 2.25B & No & 93.0 & 94.0 & 91.0 & 77.0 & 88.8  \\
& GR00T N1~\cite{bjorck2025gr00t} & 2B  & Yes & 94.4 & 97.6 & 93.0 & \underline{90.6} & 93.9  \\
& $\pi_0$~\cite{black2410pi0} & 3.5B  & Yes & \textbf{96.8} & \textbf{98.8} & \underline{95.8} & 85.2 & \underline{94.2}  \\
\midrule
& \textbf{Evo-1 (Ours)} & 0.77B  & No & 92.7 & \underline{97.7} & \textbf{96.3} & \textbf{92.3} & \textbf{94.8}  \\
\midrule
\textbf{RoboTwin} & & & & \textbf{Click Alarmclock} & \textbf{Dump Bin Bigbin} & \textbf{Place Bread Basket} & \textbf{Place Can Basket} & \textbf{Avg.}  \\
\midrule
 & & & & {\phantom{ }easy \textbar\ hard}
 & {\phantom{ }easy \textbar\ hard} & {\phantom{1 }easy \textbar\ hard} & {\phantom{1 }easy \textbar\ hard} &    \\
& ACT~\cite{zhao2023learning} & -  & No & { 32.0 \textbar\ \phantom{1}4.0}
 & { 68.0 \textbar\ \phantom{1}1.0}
 & { \phantom{1}6.0 \textbar\ 0.0}
 & { \phantom{1}1.0 \textbar\ 0.0}
 & 14.0  \\
& Diffusion Policy~\cite{chi2023diffusion} & -  & No & { 61.0 \textbar\ \phantom{1}5.0}
 & { 49.0 \textbar\ \phantom{1}0.0}
 & { 14.0 \textbar\ 0.0}
 & { 18.0 \textbar\ 0.0}
 & 18.4  \\
& RDT~\cite{liu2024rdt} & 1.2B  & Yes & { 61.0 \textbar\ \underline{12.0}}
 & { 64.0 \textbar\ \underline{32.0}}
 & { 10.0 \textbar\ 2.0}
 & { 19.0 \textbar\ \textbf{6.0}}
 & 25.8  \\
& $\pi_0$~\cite{black2410pi0} & 3.5B  & Yes & { \underline{63.0} \textbar\ 11.0}
 & { \textbf{82.0} \textbar\ 24.0}
 & { \textbf{17.0} \textbar\ \textbf{4.0}}
 & { \textbf{41.0} \textbar\ \underline{5.0}}
 & \underline{30.9}  \\
\midrule
     & \textbf{Evo-1 (Ours)} & 0.77B  & No & { \textbf{77.0} \textbar\ \textbf{58.0}}
 & { \underline{74.0} \textbar\ \textbf{37.0}}
 & { \underline{15.0} \textbar\ \underline{3.0}}
 & { \underline{37.0} \textbar\ 1.0}
 & \textbf{37.8} & \\

\bottomrule
\end{tabular}
}
\caption{\textbf{Simulation benchmark results on Meta-World, LIBERO, and RoboTwin.} 
We evaluate Evo-1 against representative baselines on three widely used simulation benchmarks. Params denotes model size (in billions); 
Robo-Pretrain shows whether the model is pretrained on robot data; 
\textbf{Bold} marks the best result, and \underline{underline} denotes the second best.}
\label{tab:combined_results}
\end{table*}

\subsection{Two-Stage Training Procedure}

To strike a balance between preserving the inherent multimodal understanding of the vision-language backbone and adapting it to downstream action generation, we adopt a two-stage training paradigm. 
Preserving the pretrained multimodal semantics is essential for maintaining the generalization ability of the model across diverse visual-linguistic contexts, preventing overfitting to specific manipulation tasks. 
At the same time, effective adaptation to action generation is necessary to ensure that the fused perceptual representations can accurately guide the diffusion-based action expert, thereby improving task success rates in downstream control. 
Direct end-to-end training would risk disrupting the pretrained representations, reducing the model’s inherent multimodal understanding and leading to overfitting on specific downstream tasks, which ultimately compromises its generalization ability.

\begin{figure}[t]
  \centering
  \begin{subfigure}[b]{0.31\linewidth}
    \centering
    \includegraphics[width=\linewidth]{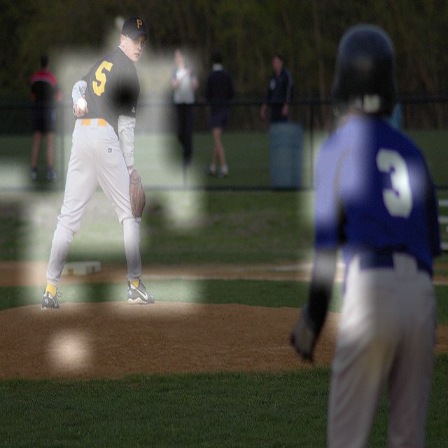}
  \end{subfigure}
  \hfill
  \begin{subfigure}[b]{0.31\linewidth}
    \centering
    \includegraphics[width=\linewidth]{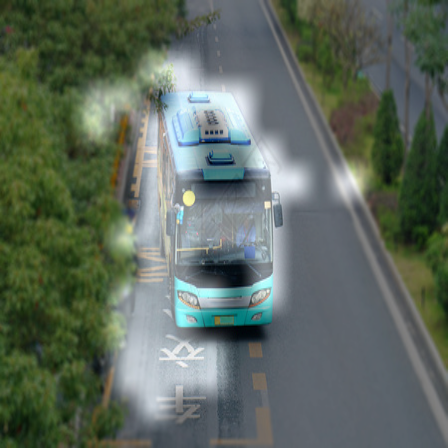}
  \end{subfigure}
  \hfill
  \begin{subfigure}[b]{0.31\linewidth}
    \centering
    \includegraphics[width=\linewidth]{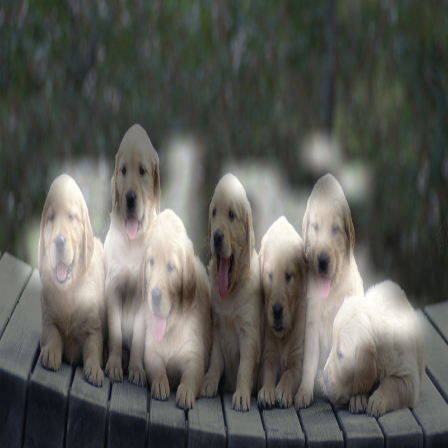}
  \end{subfigure}

  \vspace{0.2em}
  \centerline{(a) Attention maps from InternVL3-1B (ours)}

  \vspace{0.5em}

  \begin{subfigure}[b]{0.31\linewidth}
    \centering
    \includegraphics[width=\linewidth]{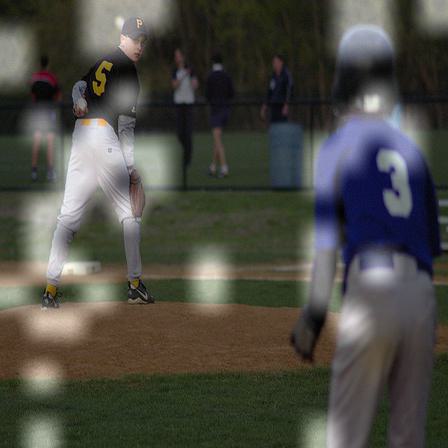}
  \end{subfigure}
  \hfill
  \begin{subfigure}[b]{0.31\linewidth}
    \centering
    \includegraphics[width=\linewidth]{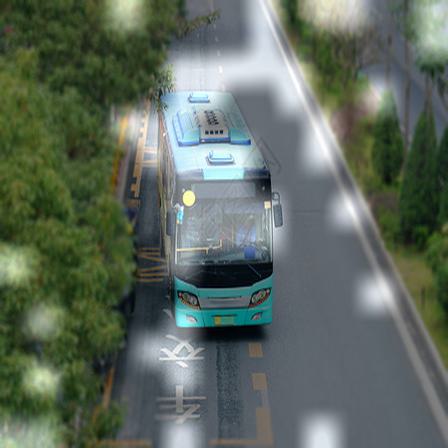}
  \end{subfigure}
  \hfill
  \begin{subfigure}[b]{0.31\linewidth}
    \centering
    \includegraphics[width=\linewidth]{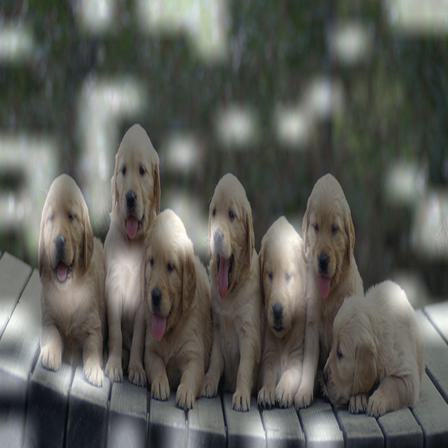}
  \end{subfigure}

  \vspace{0.5em}
  \centerline{(b) Attention maps from Prismatic-7B (OpenVLA)}

  \caption{
  \textbf{Comparison of vision-language attention maps after training.} 
(a) Evo-1 (InternVL3-1B) yields spatially consistent and semantically aligned activations.  
(b) OpenVLA (Prismatic-7B) shows degraded coherence in attention maps.
}
  \label{fig:embedding_map}
\end{figure}

\noindent\textbf{Stage 1: Action Expert Alignment.} In the first stage, we freeze the entire vision-language backbone and exclusively train the action expert along with the integration module. This setup allows the randomly initialized weights in action expert to gradually align with the multimodal embedding space without back-propagating noisy gradients into the pretrained backbone. As a result, the model can establish a coherent alignment between the VLM features and the action expert before full fine-tuning.

\noindent\textbf{Stage 2: Full-scale Fine-Tuning.} Once the integration and action module are sufficiently aligned, we unfreeze the VLM backbone and perform full-scale fine-tuning across the entire architecture. This stage enables joint refinement of both the pretrained vision-language backbone and the action expert, ensuring deeper integration and better adaptation to diverse manipulation tasks.

\noindent\textbf{Preserving Multimodal Semantics.}  
To further validate the benefit of our training strategy, we compare the image-text attention maps produced by InternVL3-1B (from Evo-1 after two-stage training) and Prismatic-7B VLM (used in OpenVLA). As illustrated in Figure~\ref{fig:embedding_map}, the embeddings from InternVL3-1B retain clearer structure and semantically consistent attention regions after training on robot manipulation data, whereas those from Prismatic-7B exhibit notable semantic drift and degraded alignment. This result shows that our training procedure effectively preserves the original semantic space, allowing the model to maintain strong visual-language understanding while adapting to downstream control tasks.

\section{Experiments}
\label{sec:experiments}

\subsection{Simulation Experiments}
\subsubsection{Meta-World Benchmark}

\noindent \textbf{Setup.} To evaluate the manipulation capabilities of Evo-1, we conduct experiments on the Meta-World benchmark~\cite{yu2020meta}. For our experiments, we generate 50 demonstrations per task, evaluate each task over ten trials, and report the average performance across five independent runs.
Following prior work~\cite{wen2025tinyvla,shukor2025smolvla}, all tasks are divided into four difficulty levels (easy, medium, hard, and very hard).
Under this standardized evaluation setup, we compare Evo-1 with several representative baselines on the Meta-World benchmark
(1) Diffusion Policy~\cite{chi2023diffusion}
(2) TinyVLA~\cite{wen2025tinyvla}
(3) $\pi_0$~\cite{black2410pi0}
(4) SmolVLA~\cite{shukor2025smolvla}.
All baseline performances are reported from their original papers or reproduction of other published works.

\noindent \textbf{Results.} As shown in Table~\ref{tab:combined_results}, Evo-1 achieves the best overall performance on the Meta-World benchmark, establishing a new state-of-the-art result among existing Vision-Language-Action models. Despite having only 0.77B parameters, Evo-1 attains an average success rate of 80.6\%, significantly surpassing much larger models such as SmolVLA (2.25B, 68.2\%) and $\pi_0$ (3.5B, 47.9\%).
Moreover, Evo-1 consistently outperforms all baselines across the four difficulty levels (easy, medium, hard, and very hard), demonstrating both superior efficiency and strong capability in diverse manipulation scenarios.

\subsubsection{LIBERO Benchmark}
\noindent \textbf{Setup.}
To further evaluate the manipulation capabilities of Evo-1, we conduct experiments on the LIBERO benchmark~\cite{liu2023libero}. The evaluation set consists of 40 tasks, which are grouped into four categories (spatial, object, goal, and long), each targeting a distinct aspect of manipulation and reasoning capability. We evaluate each task over ten trials and report the average performance across five independent runs.
Under this task setup, we compare Evo-1 against several representative VLA baselines:
(1) OpenVLA~\cite{kim2024openvla}
(2) CoT-VLA~\cite{zhao2025cot}
(3) $\pi_0$-FAST \cite{pertsch2025fast}
(4) SmolVLA~\cite{shukor2025smolvla}
(5) GR00T N1~\cite{bjorck2025gr00t}
(6) $\pi_0$\cite{black2410pi0}.
All baseline results are obtained from their original papers or official reproductions to ensure a fair and reliable comparison.

\noindent \textbf{Results.} As illustrated in Table~\ref{tab:combined_results}, Evo-1 attains an average success rate of 94.8\%, exceeding strong baselines such as $\pi_0$ (94.2\%) and SmolVLA (88.8\%). Across the four task categories (spatial, object, goal, long), Evo-1 maintains consistently strong results, with particularly high robustness on long tasks (92.3\%), where many existing VLAs exhibit notable degradation.

\subsubsection{RoboTwin Benchmark}
\noindent \textbf{Setup.}
To evaluate the ability in dual-arm manipulation, we conduct experiments on the RoboTwin Benchmark. Among them, we select four representative tasks: \textit{Click Alarmclock}, \textit{Dump Bin Bigbin}, \textit{Place Bread Basket}, and \textit{Place Can Basket}. 
Each task includes 50 demonstrations for training and 100 evaluation trials under two difficulty levels.
Under this evaluation setup, we compare Evo-1 against several representative VLA baselines:
(1) ACT \cite{zhao2023learning}
(2) Diffusion Policy \cite{chi2023diffusion}
(3) RDT \cite{liu2024rdt}
(4) $\pi_0$ \cite{black2410pi0}.
For fairness and consistency, all baseline results are reported as provided in the official RoboTwin publication~\cite{chen2025robotwin}.

\noindent \textbf{Results.}
As shown in Table~\ref{tab:combined_results}, Evo-1 achieves the highest overall performance on the RoboTwin suite, attaining an average success rate of 37.8\%, surpassing the previous SOTA model $\pi_0$ (30.9\%).
Notably, Evo-1 performs exceptionally well on the \textit{Click Alarmclock} task, demonstrating precise bimanual coordination and effective action consistency even without large-scale pretraining.
These results suggest that Evo-1, with its compact design, can still handle challenging dual-arm manipulation tasks with stable and coherent behavior.

\begin{figure}[!t]
  \centering
\setlength{\belowcaptionskip}{0pt}  
  \includegraphics[width=\linewidth]{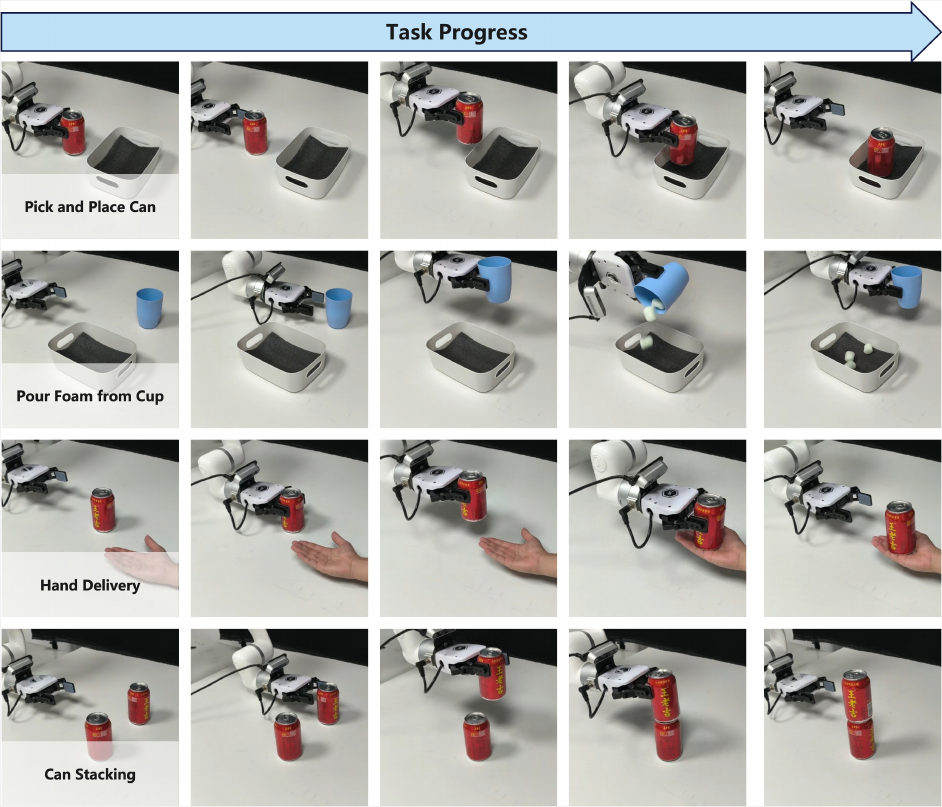}
  \caption{\textbf{Task progress of Real-World  Experiments.} Step-by-step sequences for the real-world tasks. Each row shows the detailed progression of a task from start to completion.}
  \label{fig:real_world_demo}
\end{figure}

\subsection{Real-World Experiments}

\noindent \textbf{Setup.}
To evaluate the model’s performance in diverse real-world scenarios, we conduct experiments using a 6-DoF xArm6 robotic arm equipped with a parallel gripper, and design four manipulation tasks involving diverse object manipulation and real-time interaction, as shown in Figure~\ref{fig:real_world_demo}.
\begin{enumerate} 

\item\textbf{Pick and Place Can.} This task requires grasping a beverage can from varying initial positions and placing it into a white box on the table. 

\item\textbf{Pour Foam from Cup.} This task requires lifting a foam-filled cup from varying initial positions and rotating it to pour the foam into a white box.

\item\textbf{Hand Delivery.} This task requires grasping a beverage can from varying positions and gently placing it into a human hand held at different locations.

\item\textbf{Can Stacking.} This task requires grasping a beverage can and stacking it onto another with sufficient stability. The two cans are identical and randomly placed on the table.

\end{enumerate}

For each task, we collect 100 teleoperation demonstrations to build the training dataset. Evo-1 is trained from scratch using the two-stage training process without any prior robot-data pretraining. During evaluation, each task is tested for 20 trials under varied object configurations to evaluate the stability and reliability.

\begin{figure}[!t]
  \centering
  
  \includegraphics[width=\linewidth]{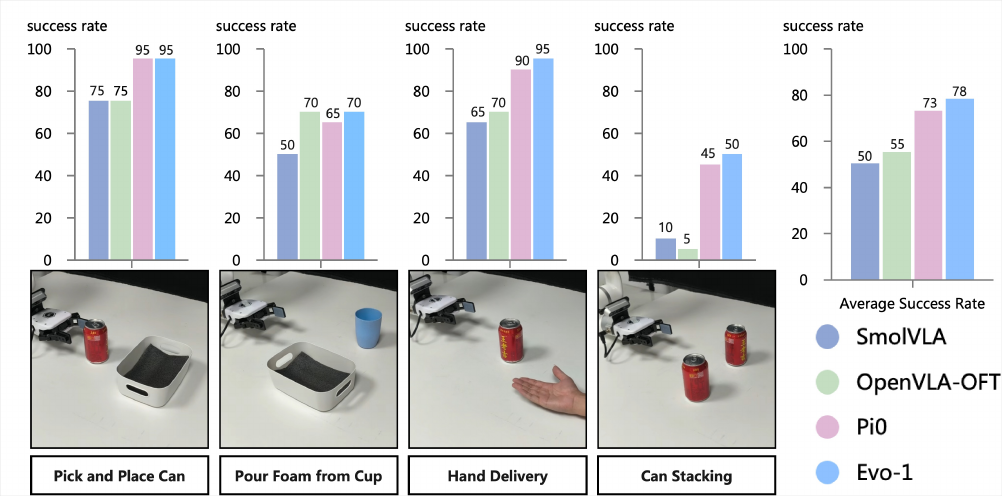} 
  \caption{\textbf{Results of Real-World experiments.} Success rates of four real-world evaluation tasks (left four subplots) and the overall average success rate across tasks (rightmost subplot).}
  \label{fig:real_world_experiments}
\end{figure}

\noindent \textbf{Results.}
As shown in Figure~\ref{fig:real_world_experiments}, Evo‑1 achieves an average success rate of 78\% across the four real-world tasks, substantially outperforming SmolVLA (50\%) and OpenVLA-OFT (55\%). With only 0.77 billion parameters (roughly one-fourth the size of 3.5-billion $\pi_0$ model), it still exceeds the performance of $\pi_0$ (73\%), highlighting its efficiency and real-world applicability.

\noindent \textbf{Inference Efficiency Analysis.}
To investigate the relationship between inference efficiency and model performance, we analyze the parameter scale, GPU memory consumption, inference frequency, and task success rate of representative VLA models in Table \ref{tab:realworld_efficiency}.
The comparison reveals a clear efficiency-performance relationship: large-scale models such as OpenVLA (7 B) and $\pi_0$ (3.5 B) require over 15 GB of GPU memory and achieve only 7-11 Hz inference frequency, while smaller models like SmolVLA (0.45 B) have lower computational demands but limited success (50 \%).
Evo-1, in contrast, strikes the best balance between efficiency and performance.
It maintains a low memory consumption of 2.3 GB, achieves the highest inference frequency of 16.4 Hz, and attains the top real-world success rate of 78\%.

\begin{table}[t]
\setlength{\tabcolsep}{3pt}
\renewcommand{\arraystretch}{0.95}

\resizebox{0.99\linewidth}{!}{ 
\begin{tabular}{lcccc}
\toprule
\textbf{Model} & \textbf{Params (B)} & \textbf{GPU Mem. (GB)} & \textbf{Infer. Freq. (Hz)} & \textbf{Success (\%)} \\ \hline
SmolVLA~\cite{shukor2025smolvla} & 0.45 & 2.0  & 12.7 & 50.0 \rule{0pt}{9pt} \\ 
OpenVLA~\cite{kim2024openvla} & 7.0 & 15.1 & 7.9 & 55.0 \\ 
$\pi_0$~\cite{black2410pi0} & 3.5 & 17.9 & 11.5 & 73.0 \\ 
Evo-1 (Ours) & 0.77 & 2.3 & 16.4 & 78.0 \\ 
\bottomrule
\end{tabular}
}
\caption{\textbf{Inference efficiency comparison.}
Comparison of model size, inference efficiency, and real-world performance on an RTX 4090d GPU.
Params (B): number of parameters (in billions);
GPU Mem.(GB): average memory usage during inference;
Infer. Freq.(Hz): average inference frequency;
Success (\%): overall success rate on real-world tasks.}
\label{tab:realworld_efficiency}
\end{table}

\subsection{Generalization Experiments}
\begin{figure}[!t]
  \centering
  \includegraphics[width=\linewidth]{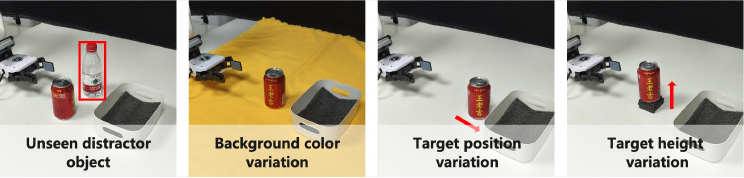}
  \caption{\textbf{Disturbance settings of generalization experiments.} We evaluate model generalization under four variations: (1) unseen distractor object, (2) background color variation, (3) target position variation, and (4) target height variation.}
  \label{fig:robustness}
\end{figure}

\noindent \textbf{Setup.}
The generalization experiments are conducted using the real-world \textit{Pick and Place Can} task as the base scenario. In each trial, the robot is required to grasp a beverage can on the table and place it into a white box. To evaluate generalization in a systematic way, we define four types of disturbance conditions, shown in Figure~\ref{fig:robustness}: (i) adding an unseen distractor object, (ii) changing the background color, (iii) shifting the target position, and (iv) varying the target height. All of these changes are beyond the training distribution. We conducted 20 trials for each disturbance condition to ensure the statistical reliability of the evaluation.

\begin{table}[]
\setlength{\tabcolsep}{5pt}
\small
\renewcommand{\arraystretch}{0.85} 
\resizebox{0.95\linewidth}{!}{ 
\begin{tabular}{lrr}
\toprule
\textbf{Condition}              & \textbf{SmolVLA} & \textbf{Ours }  \\ \hline 

\textbf{Base}                    & 75\%         & \textbf{95\%}           \rule{0pt}{10pt} \\  \hline

\textbf{Unseen distractor object} &                                  &       \rule{0pt}{10pt} \\

                     Add unseen bottle & 65\%         & \textbf{80\%}  \\ \hline
\textbf{Background color variation}                           &              &      \rule{0pt}{10pt} \\
                               Add yellow tablecloth  & 60\%         & \textbf{75\%}  \\ \hline
\textbf{Target position variation}                          &              &                \rule{0pt}{10pt} \\
                                10 mm backward       & 75\%         & \textbf{95\%}           \\
                                20 mm backward       & 60\%         & \textbf{85\%}  \\
                                30 mm backward       & 60\%         & \textbf{80\%}  \\ \hline
\textbf{Target height variation}                          &              &                \rule{0pt}{10pt} \\
                               10 mm higher         & 75\%         & \textbf{100\%}           \\
                                20 mm higher         & 65\%         & \textbf{90\%}  \\
                                30 mm higher         & 60\%          & \textbf{70\%}  \\ \bottomrule

\end{tabular}
}
\caption{\textbf{Success rates for generalization experiments.} Comparison of success rates between SmolVLA and Ours under different disturbance conditions in real-world task generalization experiments.}
\label{tab:disturbance_results}
\end{table}

\noindent \textbf{Results.}
As shown in Table~\ref{tab:disturbance_results}, Evo-1 consistently outperforms SmolVLA across all disturbance settings. It achieves 95\% in the base case and remains robust under unseen distractors (80\%) and background shifts (75\%), significantly surpassing SmolVLA (65\%, 60\%). For position variations, Evo-1 maintains high success rates (95\%, 85\%, 80\%) under increasing displacement, while SmolVLA degrades notably. Likewise, under height variations, Evo-1 retains strong performance (100\%, 90\%, 70\%), demonstrating superior generalization.

\subsection{Ablation Study}
\label{sec:ablation}
\subsubsection{Integration Module Analysis}
\begin{figure}[h]
  \centering
  \begin{subfigure}[t]{0.48\linewidth}
    \centering
    \includegraphics[width=\linewidth]{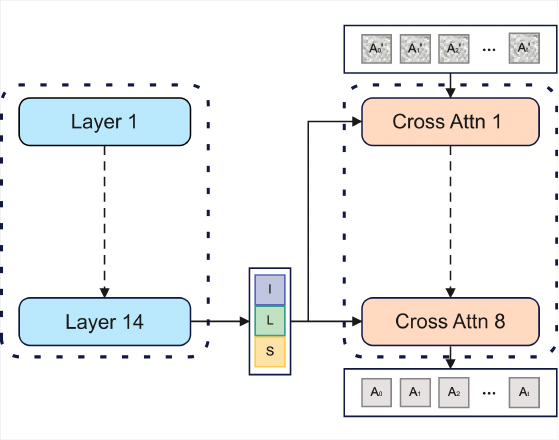}
    \caption{Module A}
  \end{subfigure}
  \hfill
  \begin{subfigure}[t]{0.48\linewidth}
    \centering
    \includegraphics[width=\linewidth]{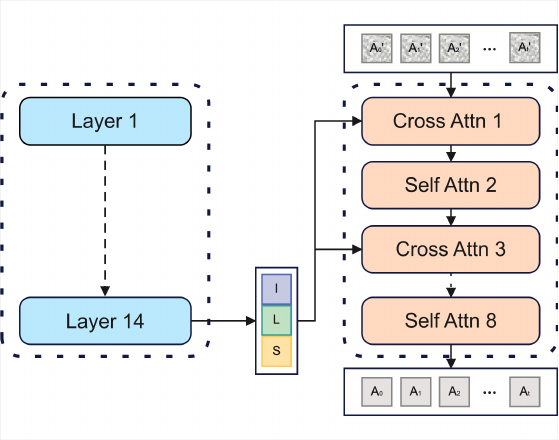}
    \caption{Module B}
  \end{subfigure}

  \vspace{0.2cm}  

  \begin{subfigure}[t]{0.48\linewidth}
    \centering
    \includegraphics[width=\linewidth]{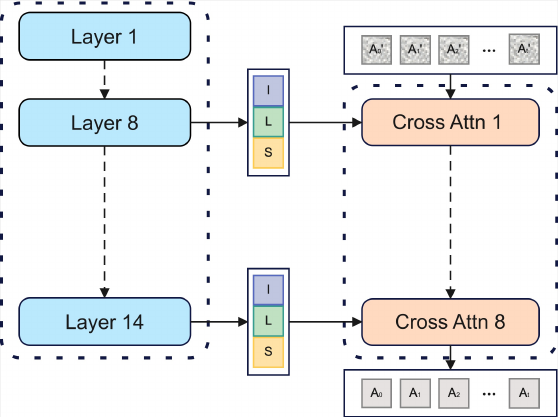}
    \caption{Module C}
  \end{subfigure}
  \hfill
  \begin{subfigure}[t]{0.48\linewidth}
    \centering
    \includegraphics[width=\linewidth]{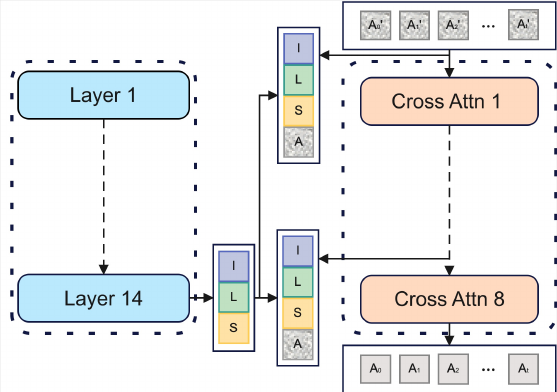}
    \caption{Module D}
  \end{subfigure}

  \caption{\textbf{Integration Module Designs.} Architectures of four different modules (A-D) for connecting the VLM and the action expert.}
  \label{fig:bridge_ablation_achitecture}
  \vspace{-8pt}
\end{figure}

We conduct experiments to investigate how different integration strategies between the Vision-Language model (VLM) and the action expert affect overall performance. 
As illustrated in Figure~\ref{fig:bridge_ablation_achitecture}, we evaluate four representative designs (Module A-D), each offering a unique approach to fusing visual, linguistic, and state information for action generation.

\noindent\textbf{Module A: Mid-Layer Cross-Attention.} This design extracts the fused multimodal feature $z_t$ from the 14\textsuperscript{th} VLM layer, concatenates it with the robot state $s_t$, and uses them as key-value inputs for all DiT layers, where the noise-injected action $A_t^{\tau}$ serves as the query in cross-attention.

\noindent\textbf{Module B: Mid-Layer Interleaved Cross-Self Attention.} This design interleaves cross-attention and self-attention layers within the DiT. Each cross-attention block attends to the concatenated VLM feature and state $s_t$, followed by a self-attention block that refines internal interactions.

\noindent\textbf{Module C: Layer-wise Cross-Attention.}
This design injects features from selected mid-to-deep VLM layers into the DiT, where each corresponding layer uses its paired VLM feature and state $s_t$ as key-value inputs, and $A_t^{\tau}$ as the query to enable hierarchical perception-action alignment.

\noindent\textbf{Module D: Joint Key-Value Cross-Attention.} This design concatenates the VLM feature, robot state, and noise-injected action to form joint key-value inputs for each DiT layer, while $A_t^{\tau}$ also serves as the query to achieve unified multimodal conditioning.

\noindent\textbf{Results.} As shown in Figure~\ref{fig:ablation_results} (a), Module A outperforms other variants by maintaining a consistent propagation of multimodal information, resulting in more coherent multimodal conditioning. In comparison, Modules~B-D introduce interruptions in this interaction process, either by inserting self-attention blocks between cross-attention layers or by using different conditioning features across layers, which breaks the continuity and consistency of information propagation.
This comparison highlights the effectiveness of Module A’s integration design, which is accordingly adopted in the final Evo-1 architecture.

\subsubsection{Training Paradigm Comparison}
We compare our proposed two-stage training paradigm with a single-stage baseline that jointly trains all modules from scratch. In the two-stage setup, we first freeze the VLM and train only the integration module and action expert. Once aligned, we unfreeze the VLM and perform full fine-tuning. In contrast, the single-stage baseline directly trains the VLM, integration module, and action expert together without any freezing schedule.

\begin{figure}[t]
  \centering
  \begin{subfigure}[b]{0.31\linewidth}
    \centering
    \includegraphics[width=\linewidth]{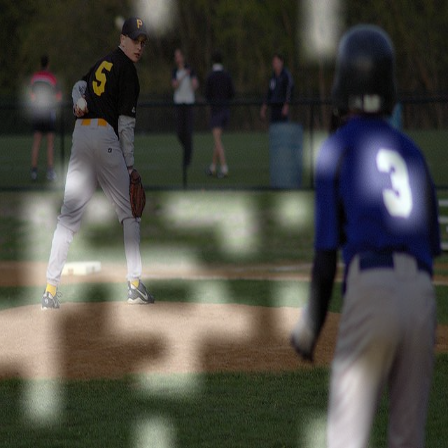}
  \end{subfigure}
  \hfill
  \begin{subfigure}[b]{0.31\linewidth}
    \centering
    \includegraphics[width=\linewidth]{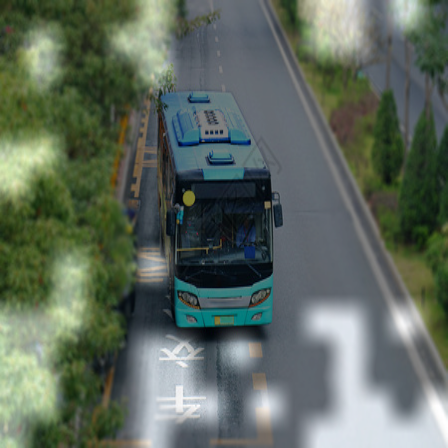}
  \end{subfigure}
  \hfill
  \begin{subfigure}[b]{0.31\linewidth}
    \centering
    \includegraphics[width=\linewidth]{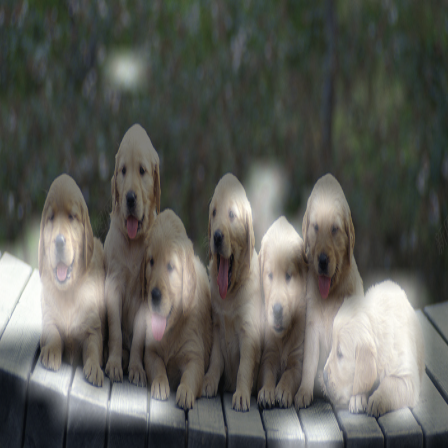}
  \end{subfigure}

  \vspace{0.2em}
  \centerline{(a) Attention maps using single-stage training paradigm}

  \vspace{0.3em}

  \begin{subfigure}[b]{0.31\linewidth}
    \centering
    \includegraphics[width=\linewidth]{pic/evo1_atten_boy.jpg}
  \end{subfigure}
  \hfill
  \begin{subfigure}[b]{0.31\linewidth}
    \centering
    \includegraphics[width=\linewidth]{pic/evo1_atten_bus.jpg}
  \end{subfigure}
  \hfill
  \begin{subfigure}[b]{0.31\linewidth}
    \centering
    \includegraphics[width=\linewidth]{pic/evo1_atten_dog.jpg}
  \end{subfigure}

  \centerline{(b) Attention maps using two-stage training paradigm (ours)}

\caption{
\textbf{Comparison of vision-language attention maps after training.} 
(a) The single-stage paradigm shows disrupted attention with reduced semantic coherence.
(b) Our two-stage paradigm preserves clear and semantically consistent focus regions.
}

  \label{fig:ablation_embedding_map}
\end{figure}

\noindent \textbf{Attention Visualization.} To analyze their difference, we visualize the attention maps of both models.
As shown in Figure~\ref{fig:ablation_embedding_map}, the two-stage paradigm preserves the semantic attention patterns of VLM, maintaining clear focus on object regions and task-relevant entities. In comparison, the single-stage training disrupts these patterns, causing the model to lose clear semantic focus and attend to irrelevant areas.

\begin{figure}[t]
\centering
\begin{subfigure}[t]{0.39\linewidth}
    \centering
    \includegraphics[height=3.6cm]{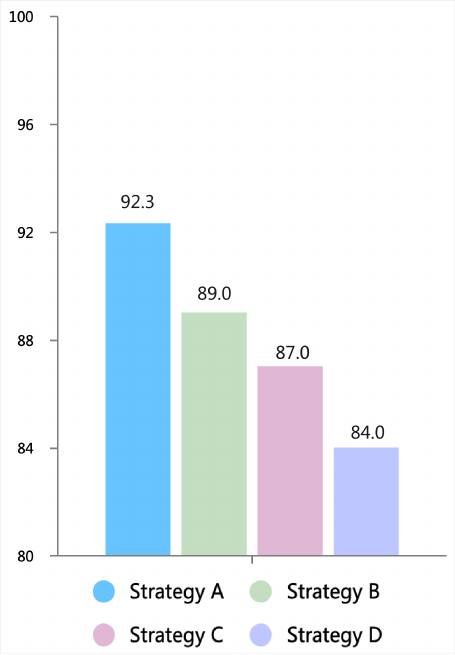}
    \caption{}
\end{subfigure}
\hfill
\begin{subfigure}[t]{0.6\linewidth}
    \centering
    \includegraphics[height=3.6cm]{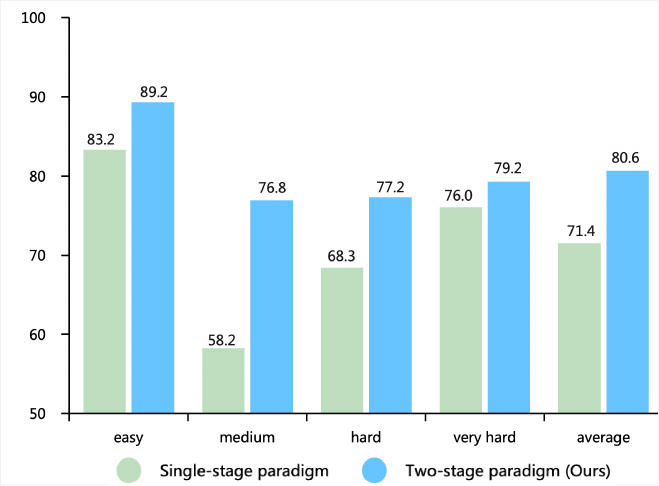}
    \caption{}
\end{subfigure}
\caption{
\textbf{Comparison results of integration modules and training paradigms.}
(a) Success rates of four integration modules on the LIBERO-Long benchmark.
(b) Performance comparison on Meta-World between a single-stage and our two-stage training paradigm.
}
\label{fig:ablation_results}
\vspace{-8pt}
\end{figure}

\noindent \textbf{Results.}
As shown in Figure~\ref{fig:ablation_results} (b), the two-stage training paradigm consistently outperforms the single-stage baseline across all difficulty levels by better preserving the perceptual representations of the vision-language backbone, thereby enhancing generalization and reducing overfitting to downstream tasks.

\section{Conclusion}
\label{sec:conclusion}
In this work, we introduce Evo-1, a lightweight and efficient Vision-Language-Action (VLA) model that enables low-cost training and high-efficiency inference on consumer-grade GPUs, while achieving state-of-the-art performance without any robot data pretraining. This achievement is attributed to our efficient architectural design and the proposed two-stage training strategy, which together ensure stable perception-action alignment while preserving the semantic understanding of vision-language backbone. 
To advance future research, we release the code, data, and model weights to encourage further research and practical development of lightweight and high performance VLA models.

{
    \small
    \bibliographystyle{ieeenat_fullname}
    \bibliography{main}

@article{alayrac2022flamingo,
  title={Flamingo: a visual language model for few-shot learning},
  author={Alayrac, Jean-Baptiste and Donahue, Jeff and Luc, Pauline and Miech, Antoine and Barr, Iain and Hasson, Yana and Lenc, Karel and Mensch, Arthur and Millican, Katherine and Reynolds, Malcolm and others},
  journal={Advances in neural information processing systems},
  volume={35},
  pages={23716--23736},
  year={2022}
}

@article{wang2024qwen2,
  title={Qwen2-vl: Enhancing vision-language model's perception of the world at any resolution},
  author={Wang, Peng and Bai, Shuai and Tan, Sinan and Wang, Shijie and Fan, Zhihao and Bai, Jinze and Chen, Keqin and Liu, Xuejing and Wang, Jialin and Ge, Wenbin and others},
  journal={arXiv preprint arXiv:2409.12191},
  year={2024}
}

@article{zhu2025internvl3,
  title={Internvl3: Exploring advanced training and test-time recipes for open-source multimodal models},
  author={Zhu, Jinguo and Wang, Weiyun and Chen, Zhe and Liu, Zhaoyang and Ye, Shenglong and Gu, Lixin and Tian, Hao and Duan, Yuchen and Su, Weijie and Shao, Jie and others},
  journal={arXiv preprint arXiv:2504.10479},
  year={2025}
}

@article{achiam2023gpt,
  title={Gpt-4 technical report},
  author={Achiam, Josh and Adler, Steven and Agarwal, Sandhini and Ahmad, Lama and Akkaya, Ilge and Aleman, Florencia Leoni and Almeida, Diogo and Altenschmidt, Janko and Altman, Sam and Anadkat, Shyamal and others},
  journal={arXiv preprint arXiv:2303.08774},
  year={2023}
}

@article{kim2024openvla,
  title={Openvla: An open-source vision-language-action model},
  author={Kim, Moo Jin and Pertsch, Karl and Karamcheti, Siddharth and Xiao, Ted and Balakrishna, Ashwin and Nair, Suraj and Rafailov, Rafael and Foster, Ethan and Lam, Grace and Sanketi, Pannag and others},
  journal={arXiv preprint arXiv:2406.09246},
  year={2024}
}

@article{black2410pi0,
  title={$\pi$0: A vision-language-action flow model for general robot control. CoRR, abs/2410.24164, 2024. doi: 10.48550},
  author={Black, Kevin and Brown, Noah and Driess, Danny and Esmail, Adnan and Equi, Michael and Finn, Chelsea and Fusai, Niccolo and Groom, Lachy and Hausman, Karol and Ichter, Brian and others},
  journal={arXiv preprint ARXIV.2410.24164}
}

@inproceedings{zitkovich2023rt,
  title={Rt-2: Vision-language-action models transfer web knowledge to robotic control},
  author={Zitkovich, Brianna and Yu, Tianhe and Xu, Sichun and Xu, Peng and Xiao, Ted and Xia, Fei and Wu, Jialin and Wohlhart, Paul and Welker, Stefan and Wahid, Ayzaan and others},
  booktitle={Conference on Robot Learning},
  pages={2165--2183},
  year={2023},
  organization={PMLR}
}

@article{lin2025evo,
  title={Evo-0: Vision-Language-Action Model with Implicit Spatial Understanding},
  author={Lin, Tao and Li, Gen and Zhong, Yilei and Zou, Yanwen and Du, Yuxin and Liu, Jiting and Gu, Encheng and Zhao, Bo},
  journal={arXiv preprint arXiv:2507.00416},
  year={2025}
}

@article{bjorck2025gr00t,
  title={Gr00t n1: An open foundation model for generalist humanoid robots},
  author={Bjorck, Johan and Casta{\~n}eda, Fernando and Cherniadev, Nikita and Da, Xingye and Ding, Runyu and Fan, Linxi and Fang, Yu and Fox, Dieter and Hu, Fengyuan and Huang, Spencer and others},
  journal={arXiv preprint arXiv:2503.14734},
  year={2025}
}

@inproceedings{yu2020meta,
  title={Meta-world: A benchmark and evaluation for multi-task and meta reinforcement learning},
  author={Yu, Tianhe and Quillen, Deirdre and He, Zhanpeng and Julian, Ryan and Hausman, Karol and Finn, Chelsea and Levine, Sergey},
  booktitle={Conference on robot learning},
  pages={1094--1100},
  year={2020},
  organization={PMLR}
}

@article{liu2023libero,
  title={Libero: Benchmarking knowledge transfer for lifelong robot learning},
  author={Liu, Bo and Zhu, Yifeng and Gao, Chongkai and Feng, Yihao and Liu, Qiang and Zhu, Yuke and Stone, Peter},
  journal={Advances in Neural Information Processing Systems},
  volume={36},
  pages={44776--44791},
  year={2023}
}

@inproceedings{o2024open,
  title={Open x-embodiment: Robotic learning datasets and rt-x models: Open x-embodiment collaboration 0},
  author={O’Neill, Abby and Rehman, Abdul and Maddukuri, Abhiram and Gupta, Abhishek and Padalkar, Abhishek and Lee, Abraham and Pooley, Acorn and Gupta, Agrim and Mandlekar, Ajay and Jain, Ajinkya and others},
  booktitle={2024 IEEE International Conference on Robotics and Automation (ICRA)},
  pages={6892--6903},
  year={2024},
  organization={IEEE}
}

@article{beyer2024paligemma,
  title={Paligemma: A versatile 3b vlm for transfer},
  author={Beyer, Lucas and Steiner, Andreas and Pinto, Andr{\'e} Susano and Kolesnikov, Alexander and Wang, Xiao and Salz, Daniel and Neumann, Maxim and Alabdulmohsin, Ibrahim and Tschannen, Michael and Bugliarello, Emanuele and others},
  journal={arXiv preprint arXiv:2407.07726},
  year={2024}
}

@article{shi2025hi,
  title={Hi robot: Open-ended instruction following with hierarchical vision-language-action models},
  author={Shi, Lucy Xiaoyang and Ichter, Brian and Equi, Michael and Ke, Liyiming and Pertsch, Karl and Vuong, Quan and Tanner, James and Walling, Anna and Wang, Haohuan and Fusai, Niccolo and others},
  journal={arXiv preprint arXiv:2502.19417},
  year={2025}
}

@article{liu2025hybridvla,
  title={Hybridvla: Collaborative diffusion and autoregression in a unified vision-language-action model},
  author={Liu, Jiaming and Chen, Hao and An, Pengju and Liu, Zhuoyang and Zhang, Renrui and Gu, Chenyang and Li, Xiaoqi and Guo, Ziyu and Chen, Sixiang and Liu, Mengzhen and others},
  journal={arXiv preprint arXiv:2503.10631},
  year={2025}
}

@article{shukor2025smolvla,
  title={Smolvla: A vision-language-action model for affordable and efficient robotics},
  author={Shukor, Mustafa and Aubakirova, Dana and Capuano, Francesco and Kooijmans, Pepijn and Palma, Steven and Zouitine, Adil and Aractingi, Michel and Pascal, Caroline and Russi, Martino and Marafioti, Andres and others},
  journal={arXiv preprint arXiv:2506.01844},
  year={2025}
}

@article{wen2025tinyvla,
  title={Tinyvla: Towards fast, data-efficient vision-language-action models for robotic manipulation},
  author={Wen, Junjie and Zhu, Yichen and Li, Jinming and Zhu, Minjie and Tang, Zhibin and Wu, Kun and Xu, Zhiyuan and Liu, Ning and Cheng, Ran and Shen, Chaomin and others},
  journal={IEEE Robotics and Automation Letters},
  year={2025},
  publisher={IEEE}
}

@article{marafioti2025smolvlm,
  title={Smolvlm: Redefining small and efficient multimodal models},
  author={Marafioti, Andr{\'e}s and Zohar, Orr and Farr{\'e}, Miquel and Noyan, Merve and Bakouch, Elie and Cuenca, Pedro and Zakka, Cyril and Allal, Loubna Ben and Lozhkov, Anton and Tazi, Nouamane and others},
  journal={arXiv preprint arXiv:2504.05299},
  year={2025}
}

@inproceedings{biderman2023pythia,
  title={Pythia: A suite for analyzing large language models across training and scaling},
  author={Biderman, Stella and Schoelkopf, Hailey and Anthony, Quentin Gregory and Bradley, Herbie and O’Brien, Kyle and Hallahan, Eric and Khan, Mohammad Aflah and Purohit, Shivanshu and Prashanth, USVSN Sai and Raff, Edward and others},
  booktitle={International Conference on Machine Learning},
  pages={2397--2430},
  year={2023},
  organization={PMLR}
}

@article{gao2024mini,
  title={Mini-internvl: a flexible-transfer pocket multi-modal model with 5\% parameters and 90\% performance},
  author={Gao, Zhangwei and Chen, Zhe and Cui, Erfei and Ren, Yiming and Wang, Weiyun and Zhu, Jinguo and Tian, Hao and Ye, Shenglong and He, Junjun and Zhu, Xizhou and others},
  journal={Visual Intelligence},
  volume={2},
  number={1},
  pages={32},
  year={2024},
  publisher={Springer}
}

@article{bai2025qwen2,
  title={Qwen2. 5-vl technical report},
  author={Bai, Shuai and Chen, Keqin and Liu, Xuejing and Wang, Jialin and Ge, Wenbin and Song, Sibo and Dang, Kai and Wang, Peng and Wang, Shijie and Tang, Jun and others},
  journal={arXiv preprint arXiv:2502.13923},
  year={2025}
}

@article{lipman2022flow,
  title={Flow matching for generative modeling},
  author={Lipman, Yaron and Chen, Ricky TQ and Ben-Hamu, Heli and Nickel, Maximilian and Le, Matt},
  journal={arXiv preprint arXiv:2210.02747},
  year={2022}
}

@article{liu2022rectified,
  title={Rectified flow: A marginal preserving approach to optimal transport},
  author={Liu, Qiang},
  journal={arXiv preprint arXiv:2209.14577},
  year={2022}
}

@inproceedings{peebles2023scalable,
  title={Scalable diffusion models with transformers},
  author={Peebles, William and Xie, Saining},
  booktitle={Proceedings of the IEEE/CVF international conference on computer vision},
  pages={4195--4205},
  year={2023}
}

@article{chi2023diffusion,
  title={Diffusion policy: Visuomotor policy learning via action diffusion},
  author={Chi, Cheng and Xu, Zhenjia and Feng, Siyuan and Cousineau, Eric and Du, Yilun and Burchfiel, Benjamin and Tedrake, Russ and Song, Shuran},
  journal={The International Journal of Robotics Research},
  pages={02783649241273668},
  year={2023},
  publisher={SAGE Publications Sage UK: London, England}
}

@article{pertsch2025fast,
  title={Fast: Efficient action tokenization for vision-language-action models},
  author={Pertsch, Karl and Stachowicz, Kyle and Ichter, Brian and Driess, Danny and Nair, Suraj and Vuong, Quan and Mees, Oier and Finn, Chelsea and Levine, Sergey},
  journal={arXiv preprint arXiv:2501.09747},
  year={2025}
}

@inproceedings{zhao2025cot,
  title={Cot-vla: Visual chain-of-thought reasoning for vision-language-action models},
  author={Zhao, Qingqing and Lu, Yao and Kim, Moo Jin and Fu, Zipeng and Zhang, Zhuoyang and Wu, Yecheng and Li, Zhaoshuo and Ma, Qianli and Han, Song and Finn, Chelsea and others},
  booktitle={Proceedings of the Computer Vision and Pattern Recognition Conference},
  pages={1702--1713},
  year={2025}
}

@article{wang2025vla,
  title={VLA-Adapter: An Effective Paradigm for Tiny-Scale Vision-Language-Action Model},
  author={Wang, Yihao and Ding, Pengxiang and Li, Lingxiao and Cui, Can and Ge, Zirui and Tong, Xinyang and Song, Wenxuan and Zhao, Han and Zhao, Wei and Hou, Pengxu and others},
  journal={arXiv preprint arXiv:2509.09372},
  year={2025}
}

@article{khazatsky2024droid,
  title={Droid: A large-scale in-the-wild robot manipulation dataset},
  author={Khazatsky, Alexander and Pertsch, Karl and Nair, Suraj and Balakrishna, Ashwin and Dasari, Sudeep and Karamcheti, Siddharth and Nasiriany, Soroush and Srirama, Mohan Kumar and Chen, Lawrence Yunliang and Ellis, Kirsty and others},
  journal={arXiv preprint arXiv:2403.12945},
  year={2024}
}

@article{zhao2023learning,
  title={Learning fine-grained bimanual manipulation with low-cost hardware},
  author={Zhao, Tony Z and Kumar, Vikash and Levine, Sergey and Finn, Chelsea},
  journal={arXiv preprint arXiv:2304.13705},
  year={2023}
}

@article{liu2024rdt,
  title={Rdt-1b: a diffusion foundation model for bimanual manipulation},
  author={Liu, Songming and Wu, Lingxuan and Li, Bangguo and Tan, Hengkai and Chen, Huayu and Wang, Zhengyi and Xu, Ke and Su, Hang and Zhu, Jun},
  journal={arXiv preprint arXiv:2410.07864},
  year={2024}
}

@article{chen2025robotwin,
  title={Robotwin 2.0: A scalable data generator and benchmark with strong domain randomization for robust bimanual robotic manipulation},
  author={Chen, Tianxing and Chen, Zanxin and Chen, Baijun and Cai, Zijian and Liu, Yibin and Li, Zixuan and Liang, Qiwei and Lin, Xianliang and Ge, Yiheng and Gu, Zhenyu and others},
  journal={arXiv preprint arXiv:2506.18088},
  year={2025}
}

@inproceedings{li2025learning,
  title={Learning precise affordances from egocentric videos for robotic manipulation},
  author={Li, Gen and Tsagkas, Nikolaos and Song, Jifei and Mon-Williams, Ruaridh and Vijayakumar, Sethu and Shao, Kun and Sevilla-Lara, Laura},
  booktitle={Proceedings of the IEEE/CVF International Conference on Computer Vision},
  pages={10581--10591},
  year={2025}
}

@article{mon2025embodied,
  title={Embodied large language models enable robots to complete complex tasks in unpredictable environments},
  author={Mon-Williams, Ruaridh and Li, Gen and Long, Ran and Du, Wenqian and Lucas, Christopher G},
  journal={Nature Machine Intelligence},
  pages={1--10},
  year={2025},
  publisher={Nature Publishing Group UK London}
}
}

\setcounter{section}{0}
\clearpage
\setcounter{page}{1}
\maketitlesupplementary

\section{Implementation Details}
\label{sec:trainingdetails}
\subsection{Training Details}
All experiments in this work are conducted with distributed training on 8$\times$~NVIDIA~A100~GPUs. 
Across all benchmarks, Evo-1 is optimized following the two-stage training paradigm introduced in the main paper, designed to ensure stable perception–action alignment while preserving the semantic structure of the pretrained VLM.

\noindent\textbf{Two-stage optimization scheme.}
The first stage focuses on stabilizing the integration layers and the action head under a fixed visual-language embedding space. 
During this stage, the backbone is entirely frozen (\texttt{finetune\_vlm = False}), while the integration module and the action expert remain trainable. 
This phase emphasizes learning a consistent alignment between vision-language representations and low-level control signals.

The second stage performs joint finetuning of all components. 
Starting from the final checkpoint of Stage 1, the backbone is unfrozen and optimized together with the integration layers and the action head. 
This stage enables Evo-1 to refine the semantic grounding of the frozen backbone and adapt it to the downstream control tasks.

\noindent\textbf{Training pipeline.}
All datasets follow the LeRobot v2.1 data format, ensuring a unified structure for images, states, and actions across different embodiments. 
Image observations are uniformly resized to $448 \times 448$. 
State and action vectors are padded to a fixed 24-dimensional representation to accommodate embodiment differences while maintaining a consistent model interface. 
For action prediction, we adopt an action trunk size of $H=50$, which serves as the horizon for the action generation process.
Data augmentation remains active throughout both stages to enhance robustness.  

In our Meta-World experiments, Stage 1 is trained for 10k steps with the VLM frozen, and Stage 2 proceeds for 65k steps with full-model finetuning enabled. Other benchmarks adopt the same overall scheme, with minor adjustments to training steps depending on dataset scale.

\subsection{Hyperparameter Details}

Tables~\ref{tab:meta_opt_hparams} and \ref{tab:meta_model_hparams} summarize the key optimization and model hyperparameters used in the Meta-World experiments, which reflect the default configuration of Evo-1 across most benchmarks. 

\begin{table}[t]
\centering
\resizebox{\columnwidth}{!}{
\begin{tabular}{lcc}
\toprule
\textbf{Hyperparameter} & \textbf{Stage 1} & \textbf{Stage 2} \\
\midrule
Learning rate & $1\times10^{-5}$ & $1\times10^{-5}$ \\
Batch size & 16 & 16 \\
Max steps & 10k & 65k \\
Warmup steps & 1k & 1k \\
Gradient clipping & 1.0 & 1.0 \\
Weight decay & 0.001 & 0.001 \\
Log interval & 10 & 10 \\
Resume from Stage 1 & No & Yes \\
\bottomrule
\end{tabular}}
\caption{Optimization hyperparameters used for Meta-World training.}
\label{tab:meta_opt_hparams}
\end{table}

\begin{table}[t]
\centering
\renewcommand{\arraystretch}{0.95} 
\setlength{\tabcolsep}{15pt}        
\resizebox{\columnwidth}{!}{
\begin{tabular}{ll}
\toprule
\textbf{Setting} & \textbf{Value} \\
\midrule
\textbf{Model Configuration} & \\
Backbone & InternVL3-1B \\
Action head & FlowMatching \\
Transformer layers & 8 \\
Dropout & 0.2 \\
\midrule
\textbf{Input Configuration} & \\
Image size & 448 \\
Use augmentation & True \\
State dimension & 24 (padded) \\
Action dimension & 24 (padded) \\
Horizon & 50 \\
\bottomrule
\end{tabular}}
\caption{Model and input configurations for Meta-World experiments.}
\label{tab:meta_model_hparams}
\end{table}

\begin{table*}[htbp]
\centering
\small
\begin{tabularx}{\textwidth}{@{}L{0.15\textwidth} L{0.28\textwidth} X@{}}
\toprule
\textbf{Difficulty} & \textbf{Task} & \textbf{Prompt} \\
\midrule

\multirow{28}{=}{\textbf{Easy}} & button-press-topdown & Press a button from the top \\
& button-press-topdown-wall & Bypass a wall and press a button from the top \\
& button-press & Press a button \\
& button-press-wall & Bypass a wall and press a button \\
& coffee-button & Push a button on the coffee machine \\
& dial-turn & Rotate a dial 180 degrees \\
& door-close & Close a door with a revolving joint \\
& door-lock & Lock the door by rotating the lock clockwise \\
& door-unlock & Unlock the door by rotating the lock counter-clockwise \\
& door & Open a door with a revolving joint \\
& drawer-close & Push and close a drawer \\
& drawer-open & Open a drawer \\
& faucet-close & Rotate the faucet clockwise \\
& faucet-open & Rotate the faucet counter-clockwise \\
& handle-press-side & Press a handle down sideways \\
& handle-press & Press a handle down \\
& handle-pull-side & Pull a handle up sideways \\
& handle-pull & Pull a handle up \\
& lever-pull & Pull a lever down 90 degrees \\
& peg-unplug-side & Unplug a peg sideways \\
& plate-slide-back-side & Get a plate from the cabinet sideways \\
& plate-slide-back & Get a plate from the cabinet \\
& plate-slide-side & Slide a plate into a cabinet sideways \\
& plate-slide & Slide a plate into a cabinet \\
& reach & Reach a goal position \\
& reach-wall & Bypass a wall and reach a goal \\
& window-close & Push and close a window \\
& window-open & Push and open a window \\
\midrule

\multirow{11}{=}{\textbf{Medium}} & basketball & Dunk the basketball into the basket \\
& bin-picking & Grasp the puck from one bin and place it into another bin \\
& box-close & Grasp the cover and close the box with it \\
& coffee-pull & Pull a mug from a coffee machine \\
& coffee-push & Push a mug under a coffee machine \\
& hammer & Hammer a screw on the wall \\
& peg-insertion-side & Insert a peg sideways \\
& push-wall & Bypass a wall and push a puck to a goal \\
& soccer & Kick a soccer into the goal \\
& sweep-into-goal & Sweep a puck into a hole \\
& sweep & Sweep a puck off the table \\
\midrule

\multirow{6}{=}{\textbf{Hard}} & hand-insert & Insert the gripper into a hole \\
& nut-assembly & Pick up a nut and place it onto a peg \\
& pick-out-of-hole & Pick up a puck from a hole \\
& pick-place & Pick and place a puck to a goal \\
& push-back & Push the puck back to a goal \\
& push & Push the puck to a goal \\
\midrule

\multirow{5}{=}{\textbf{Very Hard}} & nut-disassemble & Pick a nut out of a peg \\
& pick-place-wall & Pick a puck, bypass a wall and place the puck \\
& shelf-place & Pick and place a puck onto a shelf \\
& stick-pull & Grasp a stick and pull a box with the stick \\
& stick-push & Grasp a stick and push a box using the stick \\
\bottomrule
\end{tabularx}
\caption{Meta-World Tasks Grouped by Difficulty}
\label{tab:metaworld-difficulty}
\end{table*}

\section{Details of Simulation Experiments}
\subsection{Details of Meta-World Benchmark}
\label{sec:metaworlddetails}
\noindent\textbf{Task Setup.}
The Meta-World benchmark consists of 50 distinct robotic manipulation tasks designed for evaluating multi-task learning algorithms. For each task, the benchmark provides multiple task variations, such as different initial object positions or goal configurations, enabling the study of generalization across a broad task distribution rather than narrow parametric changes. The goal of the benchmark is to support the development of algorithms capable of acquiring new tasks more efficiently by leveraging prior experience. To this end, the collection of 50 manipulation tasks forms a sufficiently diverse task distribution intended to encourage policies to generalize to entirely new, held-out tasks.

For each of the 50 Meta-World manipulation tasks, we collect 50 high-quality demonstration trajectories. All demonstrations are obtained under the same observation and action interface described in the benchmark, with randomized initial object and goal configurations for every rollout to ensure 
sufficient intra-task variability. Across all 50 tasks, this results in a total of 2,500 demonstrations. The complete list of task names and their corresponding descriptions is provided in Table~\ref{tab:metaworld-difficulty}, which enumerates all 50 tasks used in our simulation experiments.

\begin{figure*}[!t]
  \centering
  \includegraphics[width=\textwidth]{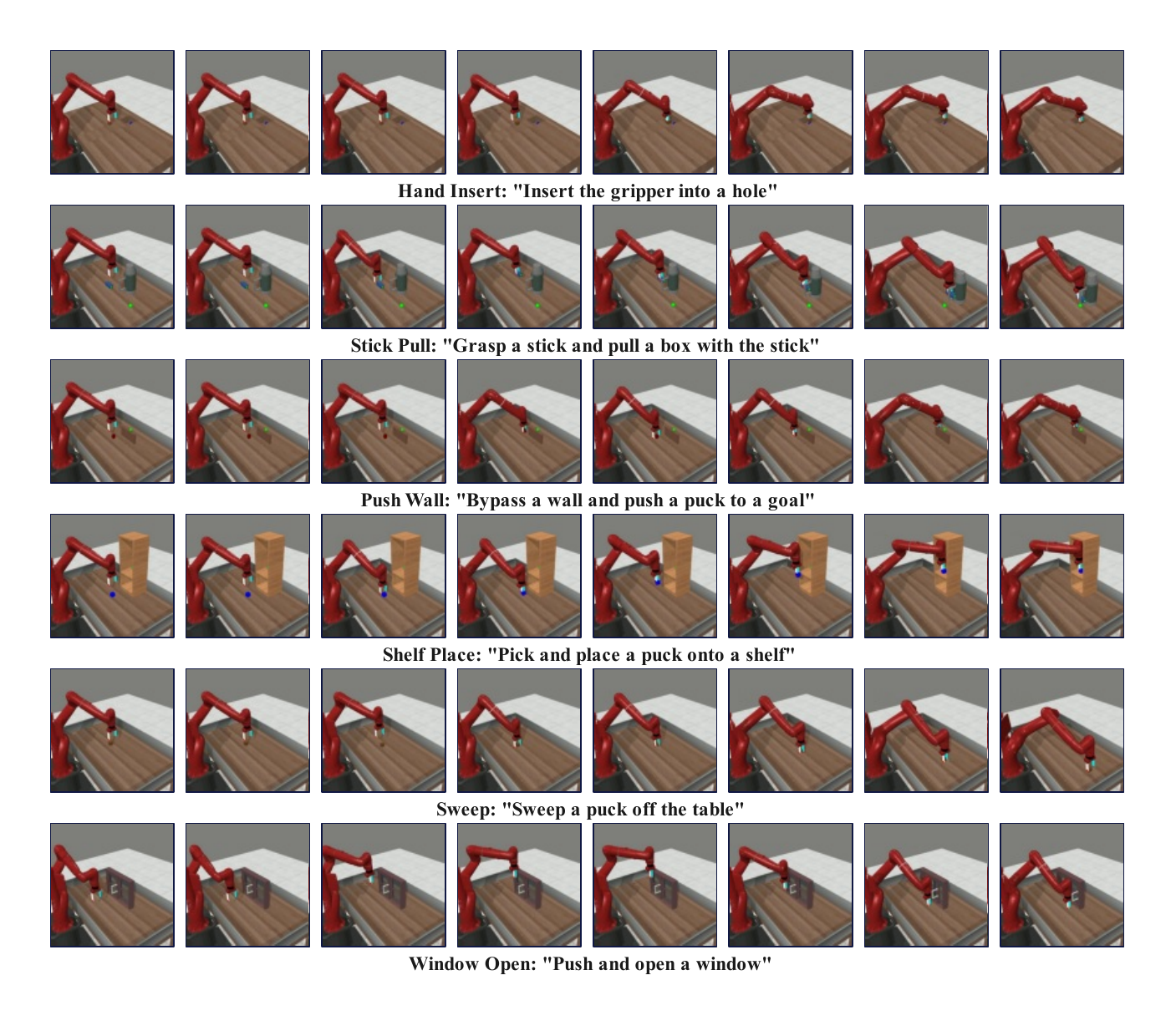} 
\caption{\textbf{Examples of Meta-World Execution.}}
  \label{fig:appendix_Metaworld}
\end{figure*}

\noindent\textbf{Meta-World Execution Examples.}
Figure~\ref{fig:appendix_Metaworld} presents several representative execution trajectories produced by our policy on different Meta-World tasks. Each example highlights the overall manipulation process, showing how the agent observes the scene, generates appropriate actions, and completes the task under varying object positions and environmental conditions.

\subsection{Details of LIBERO Benchmark}
\label{sec:LIBEROdetails}
\noindent\textbf{Task Setup.}
LIBERO is a benchmark designed to evaluate lifelong learning in robot manipulation, focusing on the transfer of both declarative and procedural knowledge. The tasks in LIBERO are categorized into four primary task suites as shown in Table~\ref{tab:libero_tasks}: LIBERO-Spatial, LIBERO-Object, LIBERO-Goal, and LIBERO-Long. Each suite aims to test different aspects of robot learning, including how well the robot can generalize learned knowledge to new situations, transfer spatial and object-specific information, and adapt to diverse task goals.
\begin{enumerate}
   
\item \textbf{LIBERO-Spatial.} This suite tests the robot’s ability to understand and manipulate spatial relationships between objects. In tasks from this suite, the robot is tasked with manipulating objects (such as placing a bowl on a plate), where the challenge lies in learning how objects relate spatially in the environment.

\item \textbf{LIBERO-Object.} Tasks in this suite test the robot’s ability to recognize and manipulate different objects. The primary challenge is in learning to handle various objects, each with different shapes, sizes, and properties.

\item \textbf{LIBERO-Goal.} These tasks focus on goal-directed manipulation where the robot must learn to achieve specific task goals, such as placing objects in predefined locations. The goal is usually fixed, but the actions to reach the goal may vary.

\item \textbf{LIBERO-Long.} This suite combines elements from the previous three, introducing more complex tasks that require a combination of declarative and procedural knowledge. The LIBERO-100 suite is more diverse, with a larger set of tasks, enabling a thorough evaluation of the robot's ability to perform a wide range of tasks and generalize across various environments. The suite includes both short-horizon tasks and long-horizon tasks, with 90 tasks requiring quick decision-making and 10 tasks designed to test long-term planning and execution.
\end{enumerate}

The goal of these tasks is not only to test the robot's ability to perform specific tasks but also to understand how well the robot can transfer learned skills across different task settings, handling new object types, new goals, and different spatial arrangements.

\begin{table*}[t]
\centering
\small
\begin{tabular}{l p{14.5cm}}
\toprule
\textbf{Task Category} & \textbf{Task Instruction} \\
\midrule

\multirow{10}{*}{LIBERO-Spatial} 
& Pick up the black bowl between the plate and the ramekin and place it on the plate. \\
& Pick up the black bowl next to the ramekin and place it on the plate. \\
& Pick up the black bowl from table center and place it on the plate. \\
& Pick up the black bowl on the cookie box and place it on the plate. \\
& Pick up the black bowl in the top drawer of the wooden cabinet and place it on the plate. \\
& Pick up the black bowl next to the ramekin and place it on the plate. \\
& Pick up the black bowl next to the cookie box and place it on the plate. \\
& Pick up the black bowl on the stove and place it on the plate. \\
& Pick up the black bowl next to the plate and place it on the plate. \\
& Pick up the black bowl on the wooden cabinet and place it on the plate. \\

\midrule

\multirow{10}{*}{LIBERO-Object} 
& Pick up the orange juice and place it in the basket. \\
& Pick up the cream cheese and place it in the basket. \\
& Pick up the salad dressing and place it in the basket. \\
& Pick up the BBQ sauce and place it in the basket. \\
& Pick up the ketchup and place it in the basket. \\
& Pick up the tomato sauce and place it in the basket. \\
& Pick up the butter and place it in the basket. \\
& Pick up the milk and place it in the basket. \\
& Pick up the chocolate pudding and place it in the basket. \\
& Pick up the orange juice and place it in the basket. \\

\midrule

\multirow{10}{*}{LIBERO-Goal} 
& Open the middle drawer of the cabinet. \\
& Put the bowl on the stove. \\
& Put the wine bottle on top of the cabinet. \\
& Open the top drawer and put the bowl inside. \\
& Put the bowl on top of the cabinet. \\
& Push the plate to the front of the stove. \\
& Put the cream cheese in the bowl. \\
& Turn on the stove. \\
& Put the black bowl on the top drawer of the cabinet. \\
& Put the wine bottle on the rack. \\

\midrule

\multirow{10}{*}{LIBERO-Long} 
& Put both the alphabet soup and the tomato sauce in the basket. \\
& Put both the cream cheese box and the butter in the basket. \\
& Turn on the stove and put the moka pot on it. \\
& Put the black bowl in the bottom drawer of the cabinet and close it. \\
& Put the white mug on the left plate and put the yellow and white mug on the right plate. \\
& Pick up the book and place it in the back compartment of the caddy. \\
& Put the white mug on the plate and put the chocolate pudding to the right of the plate. \\
& Put both the alphabet soup and the cream cheese box in the basket. \\
& Put both moka pots on the stove. \\
& Put the yellow and white mug in the microwave and close it. \\

\bottomrule
\end{tabular}
\caption{LIBERO Task Instructions}
\label{tab:libero_tasks}
\end{table*}

\begin{figure*}[!t]
  \centering
  \includegraphics[width=\textwidth]{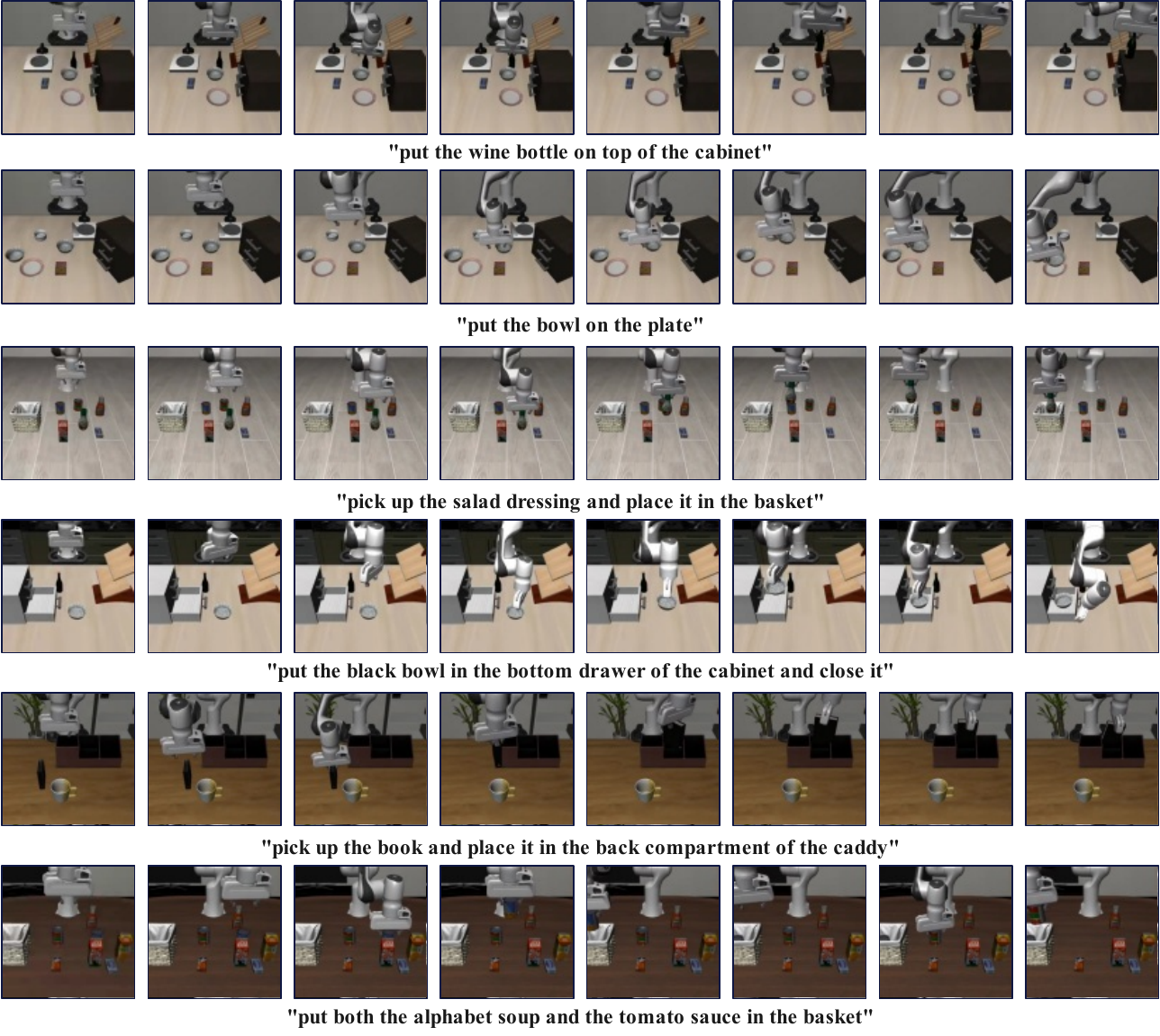} 
\caption{\textbf{Examples of LIBERO Execution.}}
  \label{fig:appendix_LIBERO}
\end{figure*}

\noindent\textbf{LIBERO Execution Examples.}
Figure~\ref{fig:appendix_LIBERO} presents qualitative examples of our policy executing several representative tasks from the LIBERO benchmark. The rollouts illustrate how the agent interprets the language instruction, identifies the relevant objects, and performs the required manipulations under different spatial arrangements and initial states. These examples highlight the model’s ability to produce reliable and consistent behaviors across a variety of LIBERO task settings.

\subsection{Details of RoboTwin Benchmark}
\label{sec:RoboTwindetails}
\noindent\textbf{Task Setup.}
To evaluate the manipulation performance of Evo-1 on dual-arm robot, we 
conduct experiments on a subset of the RoboTwin 2.0 benchmark, a diverse
manipulation suite designed to assess precision control, multi-stage
manipulation, and goal-directed object placement.  
Among the full set of RoboTwin tasks, we select four representative
manipulation scenarios:

\begin{enumerate}

    \item \textbf{Click Alarmclock.}  
    The robot is required to press the top button of an alarm clock.  
    This task evaluates fine-grained end-effector positioning and
    controlled vertical actuation necessary for successful clicking.

    \item \textbf{Dump Bin Bigbin.}  
    The robot grasps a small bin, lifts it, and dumps its contents into a
    larger bin.  
    This task requires stable grasping, lifting, rotation, and accurate
    release sequencing.

    \item \textbf{Place Bread Basket.}  
    The robot must pick up a piece of bread and place it into a basket.
    This involves handling lightweight objects and performing precise
    insertion motions into a confined receptacle.

    \item \textbf{Place Can Basket.}  
    The robot grasps a cylindrical can and places it inside a designated
    basket.  
    The task assesses grasp stability on rigid objects and accurate
    placement.
\end{enumerate}

For each of the four tasks, we collect 50 high-quality demonstration
trajectories following the RoboTwin data collection protocol.
These tasks jointly cover precision contact behaviors, multi-step motion
sequences, and constrained-object placement, enabling a comprehensive
assessment of Evo-1's manipulation generalization capabilities.

\begin{figure*}[!t]
  \centering
  \includegraphics[width=\textwidth]{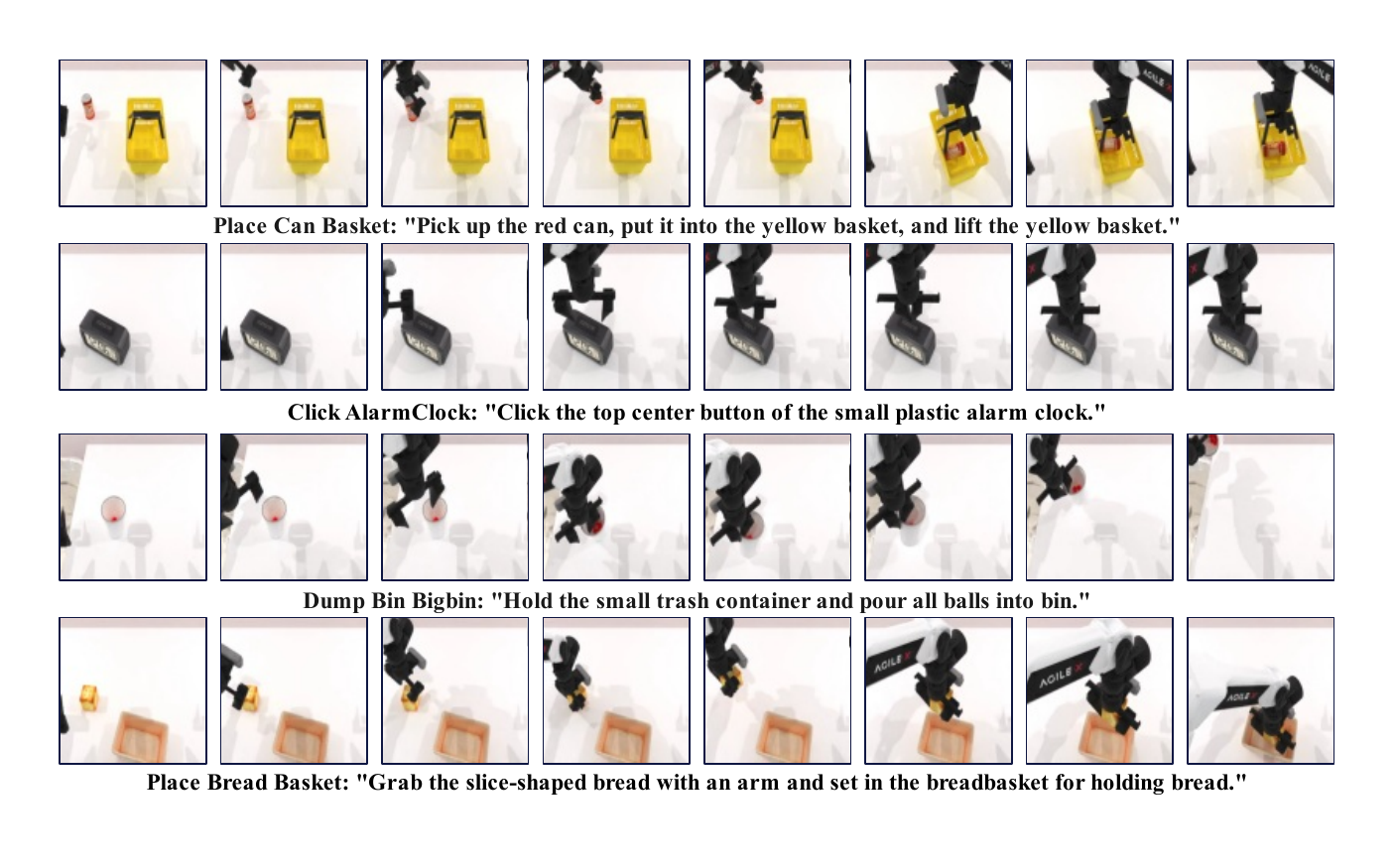} 
\caption{\textbf{Examples of RoboTwin Execution.}}
  \label{fig:appendix_Robotwin}
\end{figure*}

\noindent\textbf{RoboTwin Execution Examples.}
Figure~\ref{fig:appendix_Robotwin} illustrates several execution examples of our policy on the RobotWin benchmark. The visualized trajectories show how the agent responds to task instructions, detects the target items in cluttered scenes, and carries out the required placement or relocation actions across different object and bin configurations. These examples demonstrate that the model can maintain stable and purposeful behaviors even when facing diverse layouts and visually complex environments.

\section{Details of Real World Experiments}
\label{sec:real_world_details}
\subsection{Robot Setup}
\begin{figure}[!t]
  \centering
\setlength{\belowcaptionskip}{0pt}  
  \includegraphics[width=\linewidth]{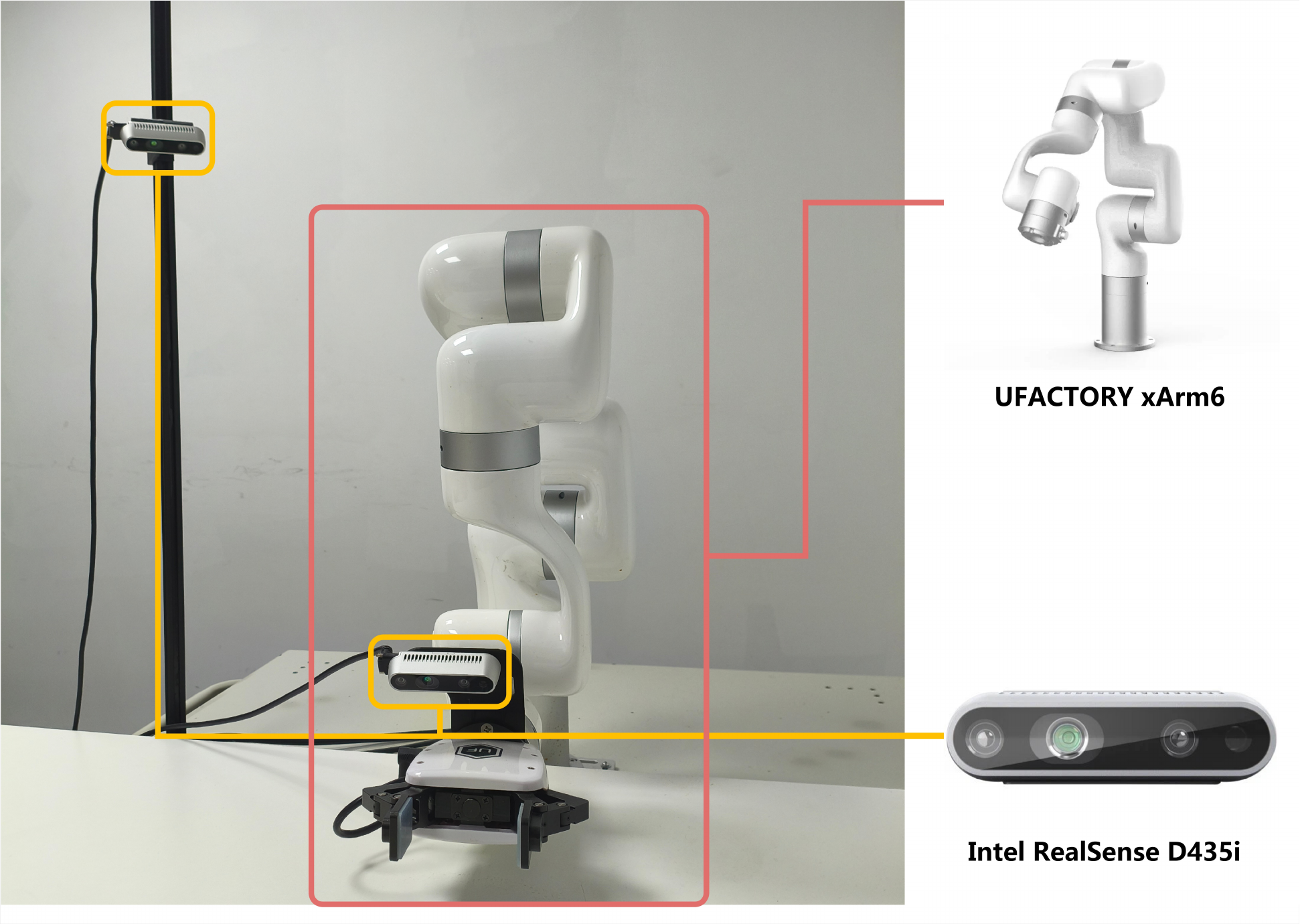}
  \caption{\textbf{Robot setup of Real-World  Experiments.} }
  \label{fig:real_world_robotsetup}
\end{figure}

As shown in Figure~\ref{fig:real_world_robotsetup}, our real-world system is built on a 6-DoF xArm6 manipulator equipped with two RGB cameras: a wrist-mounted camera that provides close-range, egocentric observations of object–gripper interactions, and a fixed environment camera that captures global scene context across the entire tabletop workspace. The robot is controlled through the official xArmAPI.

\subsection{Data Collection}
We collect real-world demonstrations following the LeRobot 2.1 data specification. Each episode consists of synchronized observations recorded at 30Hz, including two RGB image streams: \texttt{image\_1} from the fixed environment camera and \texttt{image\_2} from the wrist-mounted camera. The proprioceptive state is represented by the robot's absolute joint angles, and actions are stored using the same absolute joint angle to ensure consistent replay and supervision. All demonstrations are gathered using this unified multi-view setup, which provides both global scene context and fine-grained manipulation details that are essential for learning robust visuomotor policies.

\subsection{Task Setup and Success Criteria}

We evaluate our system on four real-world manipulation tasks. For each task, we define a clear and objective success criterion:

\begin{enumerate}
    \item \textbf{Pick and Place Can.} The robot must grasp a beverage can from varying initial positions and place it into a designated white box. A trial is considered successful if the can is fully inside the box at the end of the episode and remains stably settled without rolling out.

    \item \textbf{Pour Foam from Cup.} The robot lifts a foam-filled cup and rotates it to pour the foam into the white box. Success is achieved if the majority of the foam is correctly poured into the box such that a visible accumulation of foam is present inside, with minimal spillage outside the target area.

    \item \textbf{Hand Delivery.} The robot must grasp a beverage can and hand it over to a human operator. A trial is considered successful if the can is placed securely into the human hand, with the operator able to hold it without needing to adjust or chase the object.

    \item \textbf{Can Stacking.} The robot grasps a beverage can and stacks it onto another identical can placed on the table. Success is defined as the top can remaining stably stacked for at least two seconds without sliding or toppling.
\end{enumerate}

\subsection{Real Wolrd Execution Examples}
\begin{figure*}[!t]
  \centering
  \includegraphics[width=\textwidth]{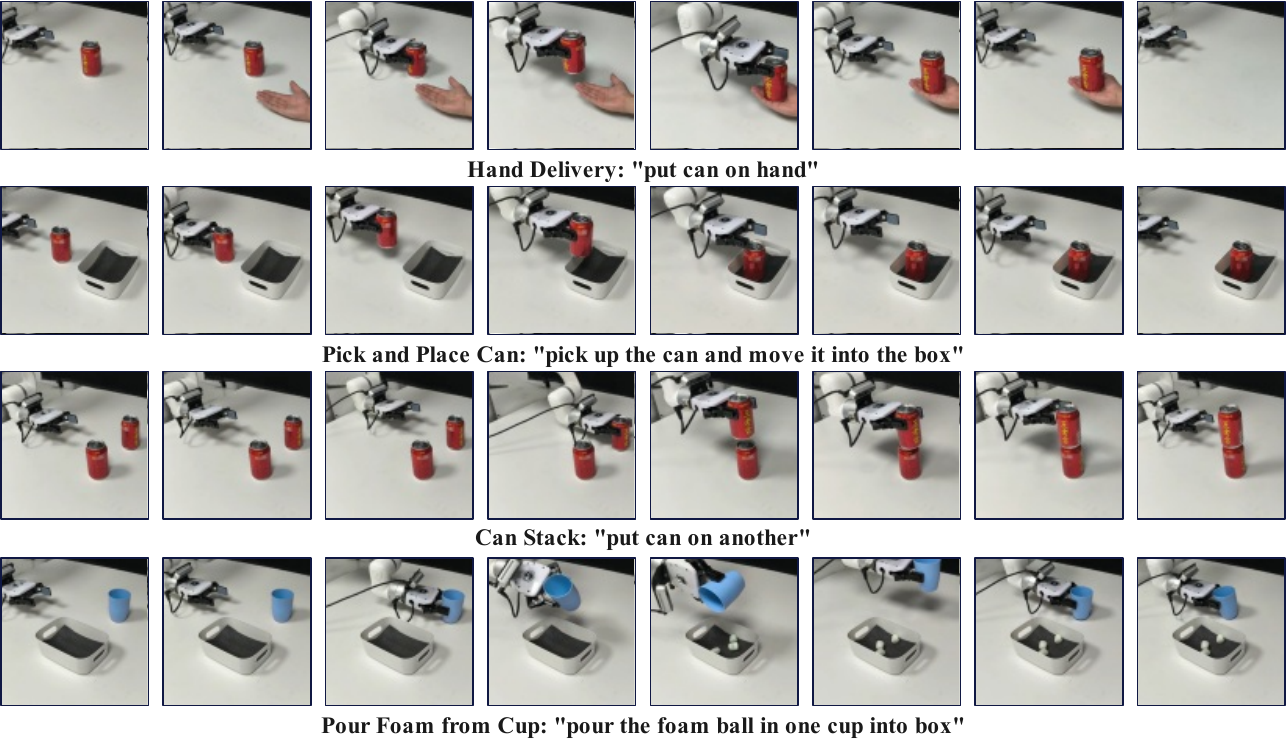} 
\caption{\textbf{Examples of Real-World Execution.}}
  \label{fig:appendix_Real}
\end{figure*}

Figure~\ref{fig:appendix_Real} provides qualitative examples of our policy performing the four real-world tasks described above. The visual sequences illustrate how the robot perceives the scene through live camera inputs, identifies the relevant objects, and executes the required manipulation behaviors under natural variations in lighting, object placement, and human interaction. Across the different task types—including container placement, pouring, handover, and stacking—the system exhibits stable motion generation and consistent task completion, demonstrating reliable transfer of the learned policy to the physical xArm6 platform.

\begin{figure}[!t]
  \centering
\setlength{\belowcaptionskip}{0pt}  
  \includegraphics[width=\linewidth]{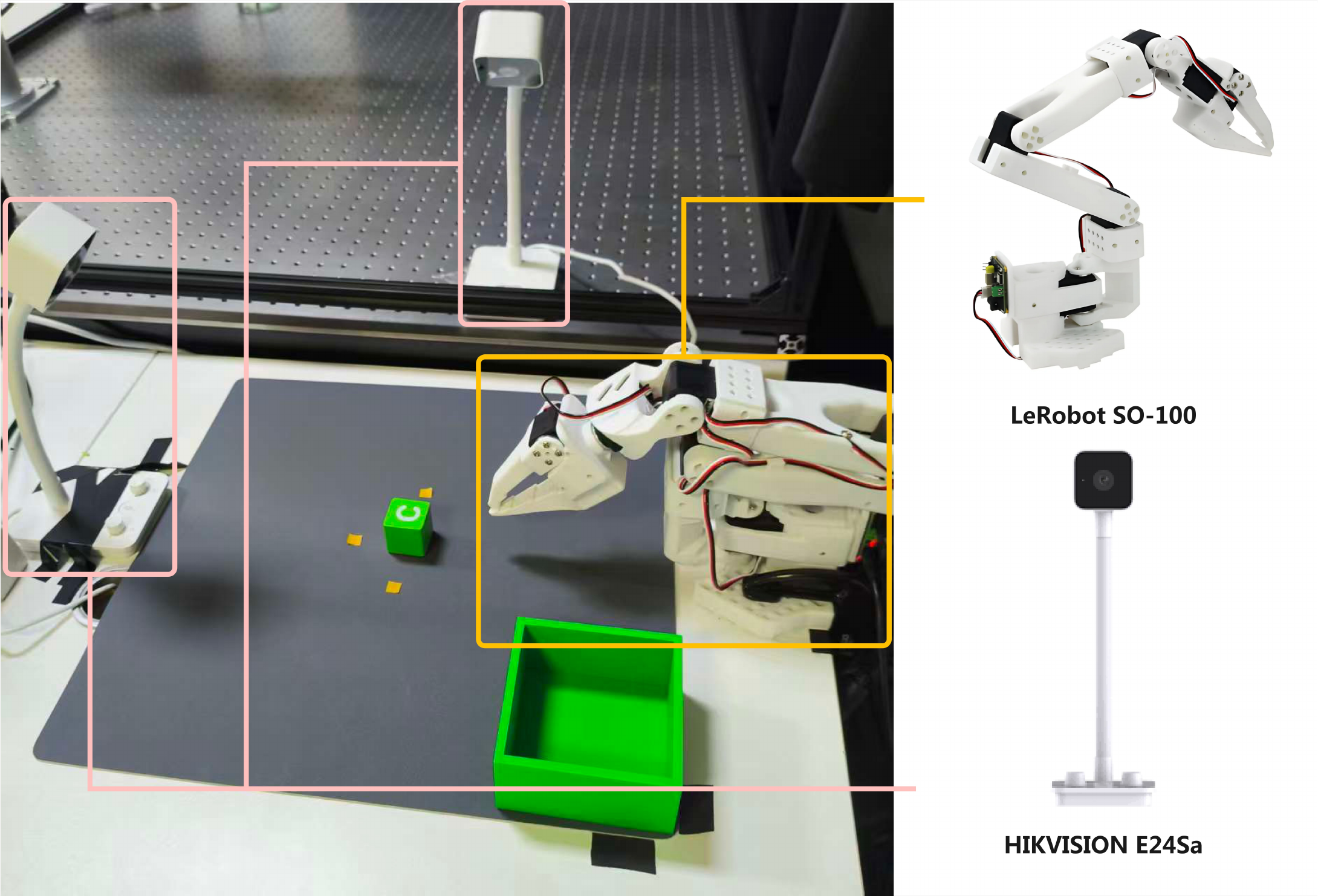}
  \caption{\textbf{Robot setup of LeRobot SO-100 Experiments.} }
  \label{fig:so100_robotsetup}
\end{figure}

\section{Deployment on LeRobot SO-100 robot}
\noindent\textbf{Task Setup.}
We deploy our policy on the LeRobot SO-100 platform, a compact desktop manipulator designed for fine-grained tabletop manipulation. In our real-world task as shown in Figure~\ref{fig:so100_robotsetup}, a small cube is manually placed at varying positions on the table. The robot must visually locate the cube, approach it with a precise grasp, lift it, and place it into a fixed container positioned on the same tabletop. This setup evaluates the system’s ability to perform reliable pick-and-place manipulation on small objects under mild variation in object pose and scene configuration.

\begin{figure*}[!t]
  \centering
  \includegraphics[width=\textwidth]{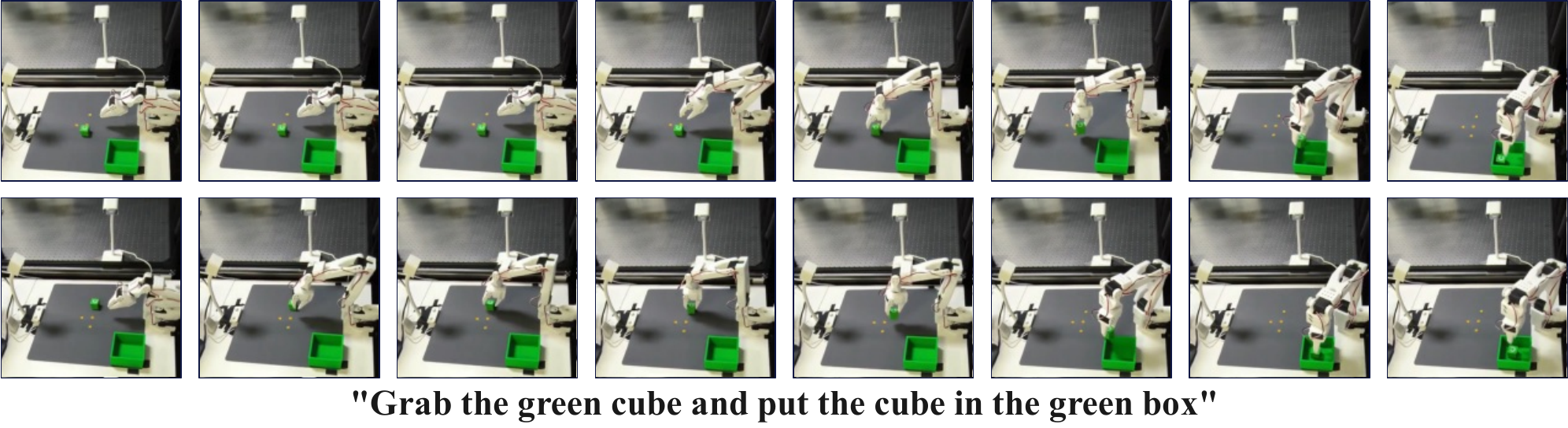} 
\caption{\textbf{Examples of SO-100 Execution.}}
  \label{fig:appendix_SO100_experiments}
\end{figure*}

\noindent\textbf{SO-100 Execution Examples.}
Figure~\ref{fig:appendix_SO100_experiments} showcases representative executions of our policy on the SO-100 desktop manipulation setup. In these examples, the robot observes the tabletop scene, identifies the target cube, and performs a smooth pick-and-place motion to deposit the object into the designated container. The rollouts illustrate that the system can handle natural variations in cube position and orientation while maintaining stable grasps and accurate placements. Overall, the results confirm that the learned policy transfers effectively to the SO-100 robot and produces reliable performance in real-world tabletop manipulation tasks.

\end{document}